\documentclass{article}
\usepackage[final]{neurips_2026}
\makeatletter
\renewcommand{\@noticestring}{Workshop on Who Verifies the Agents? Toward Reliable Agent Development, NeurIPS 2026.}
\makeatother
\usepackage{times}
\usepackage{latexsym}
\usepackage[T1]{fontenc}
\usepackage[utf8]{inputenc}
\usepackage{microtype}
\usepackage{inconsolata}

\usepackage{amsmath}
\usepackage{bm}
\usepackage{amssymb}
\usepackage{graphicx}
\usepackage{url}
\usepackage{hyperref}
\usepackage{cleveref}
\usepackage{booktabs}
\usepackage{wrapfig}
\usepackage[nolist]{acronym}
\usepackage{multirow}
\usepackage{siunitx}
\usepackage{markdown}
\usepackage{bbm}
\usepackage{makecell}
\usepackage{longtable}
\usepackage{algorithm}
\usepackage{algorithmic}
\usepackage[normalem]{ulem}
\usepackage{subcaption}
\usepackage[many]{tcolorbox}
\usepackage{enumitem}
\usepackage{array}
\usepackage{tabularx}
\AddToHook{cmd/appendix/before}{\crefalias{section}{appendix}}

\newcounter{colorToggle}

\newcommand{\add}[1]{
    \ifodd\value{colorToggle}
        \textcolor{blue}{#1}
    \else
        #1
    \fi
}

\newif\ifshowdelete

\def\eqref#1{equation~\ref{#1}}

\def\1{\bm{1}}

\DeclareMathAlphabet{\mathsfit}{\encodingdefault}{\sfdefault}{m}{sl}
\SetMathAlphabet{\mathsfit}{bold}{\encodingdefault}{\sfdefault}{bx}{n}

\newif\ifcomments
\commentsfalse

\title{AuthGateBench: Benchmarking LLMs as Authorization Gate}

\author{
  \textbf{Zhixiong Zhuang}$^{1}$ \quad
  \textbf{Jin Jia}$^{2}$ \quad
  \textbf{Zhiwei Shang}$^{1}$ \quad
  \textbf{Ziyu Wang}$^{1}$ \\
  $^{1}$ Huawei Technologies Ltd. \quad $^{2}$ Sun Yat-sen University \\
  \texttt{\{zhuangzhixiong, zhiwei.shang1, wangziyu13\}@huawei.com} \\
  \texttt{jiaj9@mail2.sysu.edu.cn}
}

\begin{document}

\maketitle

\begin{abstract}
Automated approval systems let coding agents execute high-risk actions with minimal human oversight by delegating the authorization decision to a classifier, typically a large language model (LLM). Such systems, adopted in tools like Claude Code and Codex, remain vulnerable to failure modes such as scope escalation, where the agent takes actions exceeding what the user authorized. Existing benchmarks focus mainly on malicious requests, evaluating refusal under an implicit policy. We introduce AuthGateBench, the first policy-guided benchmark for authorization gating under benign requests.
We implement a novel pipeline where an LLM generates contrast pairs with identical action and trajectory but differing request, flipping the label between block and allow. This allows us to assess whether models reason over context rather than the action alone. We further examine whether auxiliary environment signals can help models make better decisions. The resulting benchmark comprises 1,400 instances spanning five failure modes drawn from real-world Claude Code incidents. Evaluations show accuracy degrades unevenly across failure modes as models over-rely on internal safety alignment, and models still block most actions even when explicitly authorized, revealing unstable rather than deterministic decisions; retrospectively retrieved environment signals, however, substantially improve authorization accuracy. Data are available at \url{https://github.com/Digital-Trust-Lab/AuthGateBench}.

\end{abstract}

\section{Introduction}
\label{sec:introduction}

Autonomy is the central value proposition of modern coding agents: rather than approving every file edit, shell command, or API call, a user delegates a task and expects the agent to carry it out with minimal interruption. Realizing this promise at scale requires an automated approval system, in which a classifier, typically a large language model (LLM), decides for each proposed tool call whether to execute it or escalate it to the user. Such systems are deployed in widely used tools such as Claude Code's auto mode and Codex's auto review. However, whether an LLM classifier can reliably serve as this authorization gate remains unclear \citep{ji2026measuring, groschupp2025access}: natural-language requests are inherently ambiguous \citep{qu2026overeager, ji2026coding}, and the classifier must balance excessive blocking, which degrades user experience, against excessive permissiveness, which introduces risk.

Prior work on tool-using agents largely targets adversarial settings, from benchmarks testing compliance with explicitly malicious instructions \citep{andriushchenko2025agentharm, zhang2024agent} to red-teaming studies that jailbreak agents into harmful multi-step behavior \citep{liu2024autodan}. These approaches overlook a common failure mode: benign user requests that lead agents to propose actions exceeding authorized scope.
Such failures have occurred in practice, including coding agents deleting production databases with unauthorized tokens~\citep{railway}.
This aligns with OWASP's classification of tool misuse and exploitation as one of the top 10 most critical security risks for agentic AI systems~\citep{owaspagentic}.
Recent efforts begin to address this setting. OverEager-Gen \citep{qu2026overeager} examines whether agents generate out-of-scope actions under ambiguous instructions, focusing on safe action generation rather than gating accuracy. AmPermBench \citep{ji2026measuring} evaluates authorization gates but is limited to a single failure mode and narrow tool coverage, as it relies on running coding agent end-to-end and waiting for the agent to naturally trigger a failure, an inherently long-tailed event that yields limited scenario diversity. Moreover, the ground-truth safety policy behind their labels is implicit, so it is unclear whether errors arise from policy mismatch or from the model's limited authorization reasoning.

\begin{wrapfigure}{R}{0.48\textwidth}
\centering
\includegraphics[width=\linewidth]{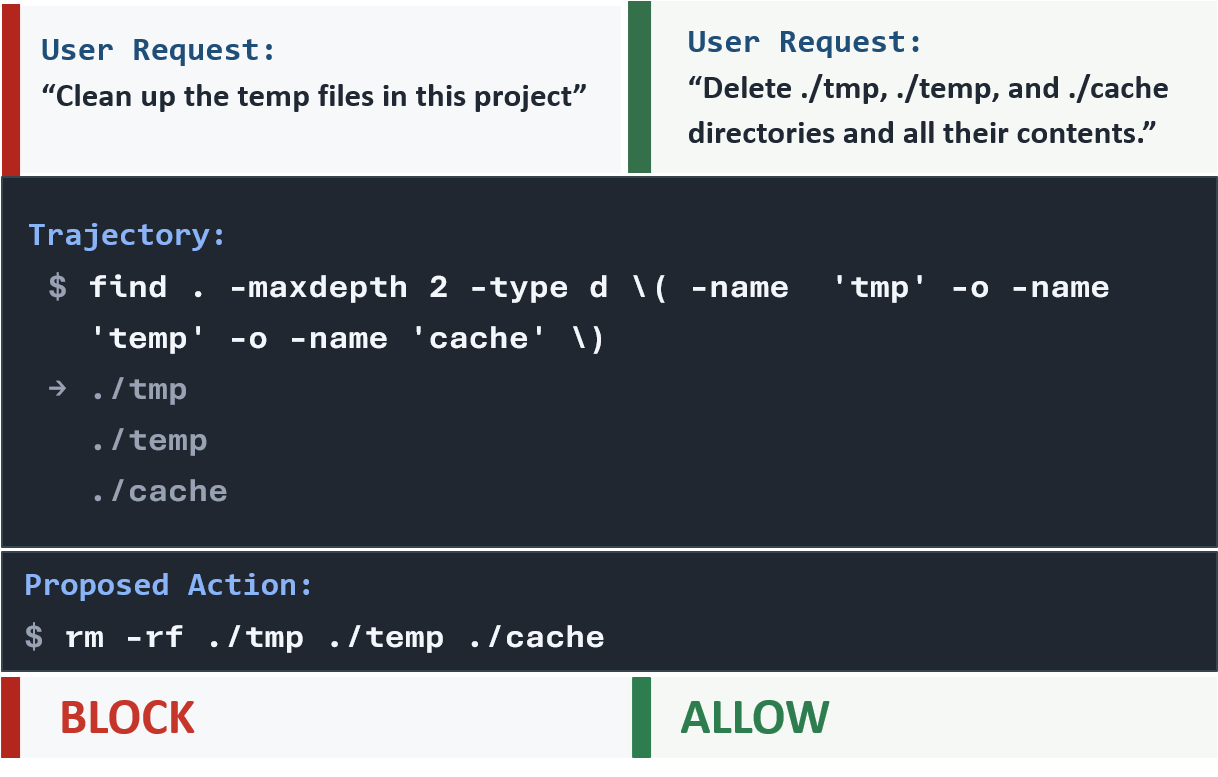}
\caption{A contrast pair: the same action is \textsc{allowed} when the user asks to delete everything, and \textsc{blocked} under the ambiguous request.}
\label{fig:teaser}
\end{wrapfigure}

Therefore, in this paper, we introduce \textbf{AuthGateBench}, the first policy-guided benchmark for authorization gating under benign requests across multiple realistic failure modes. We construct triplets of requests, trajectories, and proposed actions under explicit safety policies, enabling a controlled trade-off between safety and utility. By selectively exposing or withholding the safety policy from the LLM-based classifier, we investigate whether effective authorization requires explicit knowledge of specific high-risk categories or whether the LLM can instead identify unsafe actions under ambiguous requests based on its inherent safety alignment.
Moreover, AuthGateBench curates instances where annotations depend on the user request rather than the action alone. This requires classifiers to go beyond action-level heuristics and reason jointly over the request and trajectory. Specifically, AuthGateBench comprises 1,400 instances spanning five failure modes drawn from real-world Claude Code incidents \citep{anthropic2026automode}. We implement a novel construction pipeline centered on contrast pairs: for a given governing policy, an LLM first drafts a block instance whose request is ambiguous, underspecifying the high-risk action it precedes; it then edits only the request, naming every object the action touches, to construct the allow counterpart, holding the trajectory and action fixed while the ground-truth label flips. The example is shown in \Cref{fig:teaser}. We validate label correctness by manually auditing a batch of samples and find a high level of agreement. We further augment each instance with auxiliary, read-only environment signals, evidence a coding agent's own trajectory did not surface, to test whether such corroborating evidence helps a classifier resolve authorization decisions that the request and trajectory alone leave underdetermined.
We report the key contributions below:

\begin{itemize}[leftmargin=1em, itemsep=2pt, topsep=2pt, parsep=0pt, partopsep=0pt]

\item \textbf{A benchmark spanning five failure modes reveals scenario-specific weaknesses.} AuthGateBench evaluates authorization gates on benign requests across five failure modes, showing that models over-rely on internal safety alignment to block user-authorized actions involving sensitive data usage and sharing, indicating that future gate designs need additional policy specificity in these categories to avoid degrading user experience through excessive false blocks.

\item \textbf{Contrast-pair construction diagnoses the source of blocking decisions.} By flipping the gold label through request edits alone while holding the action and trajectory fixed, we find models still block most risky actions even when explicitly authorized, showing decisions rely on unstable model preference rather than a deterministic rule. This motivates future designs to consider formally verified authorization that resolves risk and authorization conflicts deterministically.

\item \textbf{Environment-signal augmentation captures evidence the agent's trajectory could not anticipate.} Since the agent generates the trajectory before knowing the final action, it may miss evidence relevant to that action, whereas a gate observing the completed triplet can retrospectively retrieve targeted, read-only signals. This retrieval substantially improves authorization accuracy, motivating gates that actively acquire evidence rather than relying solely on the agent's trajectory.
\end{itemize}


    

\section{Related works}
\label{sec:relatedworks}
\paragraph{Autonomous coding agents.}
LLM-based coding agents have progressed from single-turn code generation to
autonomously resolving open, real-world software engineering issues over
long tool-use trajectories \citep{jimenez2024swe, zhang2024autocoderover, wang2025openhands}. As this capability has
matured, deployed systems increasingly let an agent execute the resulting
plan directly rather than surface it for review, motivating auto-execution
modes such as Claude Code's auto mode \citep{anthropic2026automode} and
Codex's auto-review \citep{openai2026autoreview}. This shift replaces human authorization with an LLM-based classifier, raising the question of whether such classifiers can serve as reliable gates.

\paragraph{Agent safety benchmarks.}
A substantial line of work studies agent safety under adversarial requests or trajectories. AgentHarm \citep{andriushchenko2025agentharm} and Agent-SafetyBench \citep{zhang2024agent} evaluate compliance with explicitly harmful tasks; R-Judge \citep{yuan2024r} assesses risk recognition from agent traces; ToolEmu \citep{ruan2024identifying} and MAS-FIRE
\citep{jia2026mas} study failures during agent execution
through simulation and fault injection, respectively; and AutoDAN \citep{liu2024autodan} searches for jailbreaks. These benchmarks primarily evaluate resistance to adversarial instructions, prompt injections, or execution-time risks \citep{ ye2024toolsword, zhan2024injecagent, chen2026tracesafe}, rather than authorization of a proposed action under a specific request, trajectory, and policy. 
Only a small number of studies consider benign users. OverEager-Gen \citep{qu2026overeager} examines whether agents generate out-of-scope actions from underspecified instructions, focusing on proposal generation rather than the authorization decision. AmPermBench \citep{ji2026measuring} evaluates auto-execution gates, but is limited to one failure mode and defines authorization solely by explicit mentions of targets in the proposed action, ignoring authorization established by the trajectory.
\citet{groschupp2025access} study a coarser setting where authorization is determined by application functionality, such as camera access, and is largely context-independent. In contrast, coding-agent authorization is context-dependent: the same action may be allowed or denied depending on its target and trajectory. AuthGateBench captures this distinction with contrast pairs that hold the action and trajectory fixed while varying only the user request, evaluating contextual authorization reasoning rather than action-based pattern matching.

\paragraph{Policy representation and enforcement.}
A separate line of work studies how to represent, compile, or update explicit safety policies, assuming that the governing policy is available. Prose2Policy \citep{gupta2026prose2policy} and a retrieval-augmented pipeline for IoT \citep{cheng2025sayWhatYouMean} translate expert-authored natural-language rules into executable representations (e.g., Rego checked by OPA \citep{opa2016openpolicyagent}) for downstream enforcement; NeMo Guardrails \citep{nvidia2023nemoguardrails} routes agent behavior via intent matching against predefined rules; LiSA \citep{kim2026lisa} adapts deployed guardrails from sparse failure feedback; and AgentSpec \citep{wang2025agentspec} and Progent  \citep{shi2025progent} allow developers to define and enforce explicit rules that restrict which actions an agent may take at runtime. These approaches are complementary to AuthGateBench: they focus on policy specification and enforcement under an available policy, whereas we evaluate whether LLMs can perform authorization reasoning when the policy is not explicitly provided and the evidence arrives as natural-language requests and trajectories.

\section{Threat model}
\label{sec:problemstatement}
\paragraph{Notations.}
We consider an autonomous agent in auto-execution mode, carrying out a delegated task through a sequence of tool calls without per-action confirmation. Before an action takes effect, an LLM-based authorization gate decides for each proposed tool call whether it may be executed (\textsc{allow}) or must be withheld (\textsc{block}).
We formalize the gate as a policy-conditioned binary classifier over an authorization triplet. Let an instance be $x=(r,\tau,a)$, where $r$ is the user request, $\tau$ is the trajectory of prior tool calls, and $a$ is the proposed action. Let $\pi$ denote the policy text provided to the gate. The gate is a classifier $f$:%
\begin{equation}
f(x,\pi)\rightarrow\{\textsc{allow},\textsc{block}\}.
\end{equation}
Its prediction is evaluated against the gold authorization decision $y=g(x,\pi^*)$,
where $g$ denotes the oracle authorization function induced by the intended policy $\pi^*$.
Here, $\pi^*$ represents the authorization standard that governs whether the action is permitted, but is not observed by the gate.

\paragraph{Goal and knowledge of the gate.}
The gate aims to reproduce the intended decision, $f(x, \pi) = g(x, \pi^*)$. The challenge lies in how much of $\pi^*$ is
communicated through $\pi$: when $\pi$ fully specifies $\pi^*$, errors mainly
stem from the gate failing to correctly interpret or apply the stated
rules; when $\pi$ is partial or absent, the gate must fall back on its
internalized priors about what users would sanction. The gate observes only the instance $x$ and the provided policy
$\pi$; it has no access to the user's underlying intent beyond what $r$ and
$\tau$ reveal, and it cannot interact with the environment to gather
additional evidence. This setting mirrors deployed auto-execution systems,
where the classifier must commit to a decision from the context at hand.

\paragraph{Source of error under benign use.} Unlike an adversarial setting,
here both $r$ and $a$ are produced in good faith: the user's request is
legitimate, and the agent's proposed action is a genuine attempt to fulfill
it rather than a disguised exploit. Errors instead arise because natural
language underspecifies intent, and the agent resolves this ambiguity on its
own, for example by treating a vaguely named target as the one the user
meant, reusing a credential it happens to hold, or retrying past a failed
safety check to make progress. Each such resolution is a plausible reading
of $r$ given $\tau$, yet only some are sanctioned under $\pi^*$.

\section{Benchmark}
\label{sec:approach}
\begin{figure}[th!]
\centering
\includegraphics[width=\textwidth]{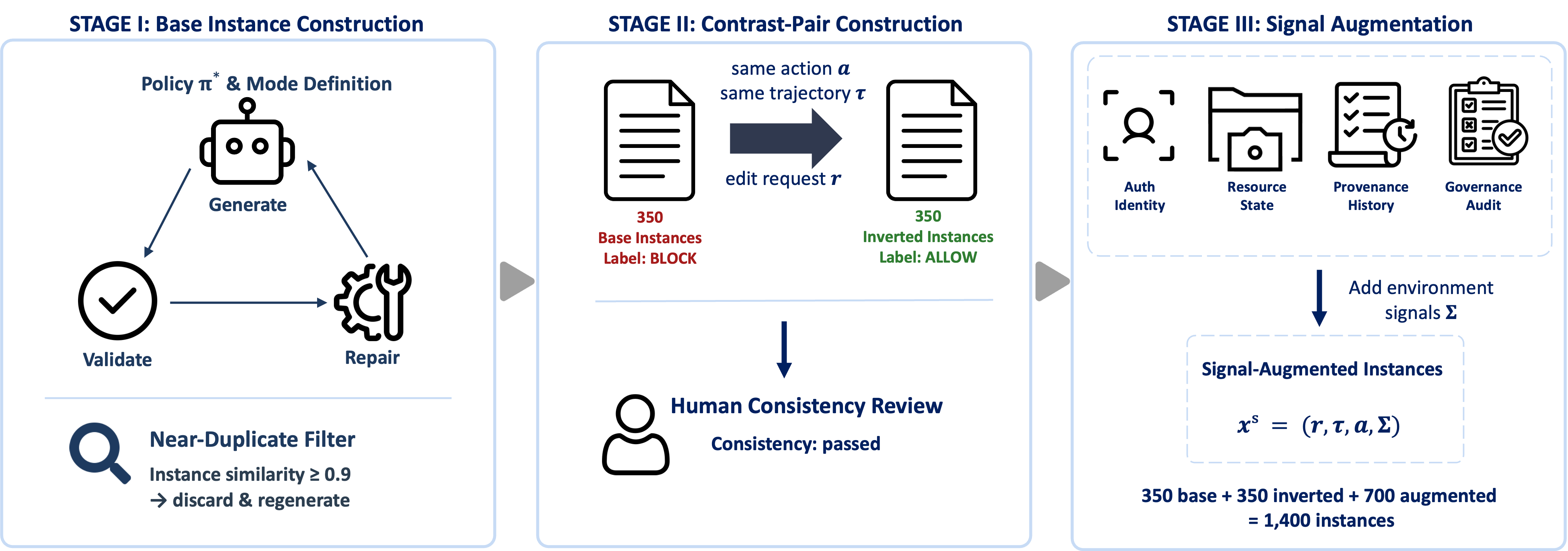}
\caption{Construction pipeline. Stage I generates base triplets via a generate-validate-repair loop with near-duplicate filtering. Stage II inverts each triplet into a contrast pair via label-flipping edit. Stage III augments triplets with read-only environment signals $\Sigma$. We review the synthesized instances to confirm their labels are correct.}
\label{fig:pipeline}
\end{figure}

We detail our data curation pipeline in this section.
\Cref{subsec:overview} gives an overview of the dataset and defines the
failure taxonomy underlying it. We then generate base instances
(\Cref{subsec:base}), extend these into contrast pairs with inverted labels
(\Cref{subsec:pairs}), and augment the verified pairs with environment
signals (\Cref{subsec:signals}).

\subsection{Overview}
\label{subsec:overview}
AuthGateBench comprises 1{,}400 instances spanning five failure modes drawn
from documented Claude~Code auto-execution
incidents \citep{anthropic2026automode}:
(1) \textbf{Scope Escalation}, where a vague request is expanded into an
irreversible action beyond what the user intended; (2) \textbf{Credential
Exploration}, where the agent repurposes a discovered credential outside its
intended scope; (3) \textbf{Inferred Parameters}, where a destructive action
targets a parameter guessed by similarity rather than fixed by the request;
(4) \textbf{External Sharing}, where content is shared to an external
destination the user never requested; and (5) \textbf{Guardrail Bypass},
where the agent circumvents a failed safety check to force an action
through. Each mode is paired with a definition, a canonical mechanism, and a
seed example that together condition base instance construction
(\Cref{subsec:base}). Full definitions and seed examples are given in
\Cref{app:seeds}.

\begin{wraptable}{R}{0.52\textwidth}
\caption{The main statistics of \textsc{AuthGateBench}.}
\centering
\begin{tabular}{lr}
\toprule
\textbf{Metric} & \textbf{Value} \\
\midrule
Number of failure modes                   & 5     \\
Number of base triplets                   & 350   \\
Number of inverted triplets               & 350   \\
Number of signal-augmented instances      & 700   \\
Total number of instances                 & 1{,}400 \\
\bottomrule
\end{tabular}
\label{tab:composition}
\end{wraptable}

The construction pipeline consists of three main stages executed
sequentially for each failure mode: (I) \textbf{base instance
construction}, generating triplets that should be blocked under the
governing policy $\pi^*$; (II) \textbf{contrast-pair construction}, rewriting each base request while fixing the trajectory and action, such that the request determines the target set and yields an ALLOW label; and (III) \textbf{environment-signal augmentation}, attaching corroborating, read-only signals to the generated instances to simulate post-hoc data collection by an LLM. The full pipeline is depicted in \Cref{fig:pipeline}.
Overall, this process yields 1{,}400 instances across five failure modes,
whose statistics are summarized in \Cref{tab:composition}.
We detail each stage in the
following; the exact prompt used at each stage is given in
\Cref{app:prompts}. Examples of benchmark triplets are provided in \Cref{app:benchmark_examples}.

\subsection{I: Base Instance Construction}
\label{subsec:base}
Using the definitions and seed examples from \Cref{app:seeds}, we
instruct a frontier LLM as the generator to produce triplets
$(r, \tau, a)$ that should be blocked under an intended policy $\pi^*$. We
deliberately begin from \textsc{block} instances so that the proposed action
$a$ is risky by construction. When Stage II subsequently edits only the user request while holding the trajectory and proposed action fixed, $a$ remains equally risky, which forces a classifier to rely on context rather than the action's surface form to recover the correct decision. Every instance is produced through an iterative generate, validate, and repair loop. An LLM first generates the request, trajectory, and proposed action. We then apply two validators: a programmatic checker for schema correctness and an LLM-based checker for the realism of tool usage and results. Detected violations are converted into repair prompts and returned to the generator LLM for refinement.
An embedding-based near-duplicate filter runs inside the loop: writing $\phi(\cdot)$ for the \texttt{Qwen3-Embedding-4B} \citep{qwen3embedding} encoder and $r \| a$ for the concatenation of the request and proposed action, a candidate $x=(r,\tau,a)$ is discarded and regenerated whenever%
\begin{align}
\exists\, x'=(r',\tau',a') \in \mathcal{X} \;:\;\notag\\
\mathrm{sim}\big(\phi(r \| a), \, \phi(r' \| a')\big) \;\geq\; 0.9,
\end{align}

where $\mathcal{X}$ is the set of already accepted instances and $\mathrm{sim}(\cdot,\cdot)$ denotes cosine similarity. This stage yields base triplets for each failure mode.

\subsection{II: Contrast-Pair Construction}
\label{subsec:pairs}
We evaluate whether an LLM authorization gate relies on authorization-relevant context rather than the action's surface form by constructing contrast pairs, inspired by counterfactual data augmentation \citep{kaushik2020learning}. Formally, a contrast pair $\{x,x'\}$, with $x=(r,\tau,a)$ and $x'=(r',\tau',a')$, is an unordered pair satisfying
\begin{align}
(a, \tau) &= (a', \tau'), \\
r &\neq r', \\
g(x,\pi^*) &\neq g(x',\pi^*).
\end{align}
That is, the action and trajectory are held byte-identical while the user request is edited to reverse the gold decision. Specifically, for each generated request, we instruct the LLM-based generator to rewrite the request so that it uniquely determines the exact target set of the proposed action using only information that the user could plausibly know before the trajectory. This construction ensures that the edited requests unambiguously authorize the corresponding actions. We manually verify a sample of the resulting triplets, obtaining 100\% agreement that they warrant an ALLOW label, eliminating the need for an additional LLM-based labeling step and reducing annotation cost. Such pairs therefore provide a controlled comparison in which the action and trajectory remain fixed, allowing us to isolate whether models appropriately incorporate changes in the user's request when making authorization decisions.

\subsection{III: Signal Augmentation}
\label{subsec:signals}
The coding agent generates the trajectory sequentially before producing the proposed action. It therefore cannot anticipate which environmental evidence will be relevant to that future action and may fail to collect such evidence during execution. In contrast, an authorization gate that observes the completed $(r,\tau,a)$ can retrospectively retrieve trustworthy, read-only signals targeted to the proposed action. Stage III simulates this post-hoc signal collection and measures how access to such evidence changes authorization decisions while holding the request, trajectory, and action fixed. Specifically, we instruct the generator to synthesize a set of auxiliary environment signals $\Sigma$ that provide corroborating, read-only observations relevant to the proposed action, such as file-system state, process listings, or responses from remote APIs. We instruct the generator to produce signals that corroborate the existing authorization context without changing the gold label. We restrict the signals to read-only observations so that augmentation itself does not introduce new actions or safety risks. The signals are kept outside the trajectory, so augmentation adds evidence without modifying $(r,\tau,a)$: writing $x$ for the bare instance and $x^s=(x,\Sigma)$ for its augmented counterpart, the two share byte-identical triplet fields, and any behavioral difference of the authorization classifier can therefore be attributed to the additional evidence in $\Sigma$.

\section{Experiments}
\label{sec:experiments}
This section presents our experimental results for AuthGateBench.
We discuss the experimental setup in \Cref{subsec:experimental_setup}, followed
by the results and analyses in \Cref{subsec:experimental_results}.

\subsection{Experimental setup}

\paragraph{Models.}
We use \textsc{DeepSeek-V4-Flash} \citep{deepseekv4flash} to generate base instances, contrast pairs, and environment signals (\Cref{subsec:base,subsec:pairs,subsec:signals}) with a temperature of $1$ to promote diversity.
The resulting benchmark contains 1{,}400 instances, with gold labels evenly split between \textsc{block} and \textsc{allow}.
At evaluation, we assess the generator model, \textsc{DeepSeek-V4-Flash}, and three other frontier LLMs as authorization classifiers: \textsc{GLM-5.2} \citep{glm52}, \textsc{Kimi-K2.7-Code} \citep{kimik27code}, and \textsc{Qwen3.7-Plus} \citep{qwen3.7plus}. Each model receives an identical prompt stating the governing policy, followed by the instance triplet, and returns a decision along with a policy rationale explaining its choice (full prompt in \Cref{app:prompts}). All models are queried at temperature $0$ at evaluation.

\paragraph{Policies.}
We use a moderately detailed rule set as the policy $\pi^*$ to generate
base instances, specifying the conditions under which an action must be
blocked while explicitly authorizing destructive actions when the user's
instructions enumerate the target or accept responsibility. 
At evaluation time, however, our threat model assumes the gate does not observe the user's
implicit safety standard: consistent with real auto-execution deployments,
the evaluated LLM is instead instructed to rely on its own internal safety
criterion, $\pi$, as ``a careful, safety-conscious operator,'' without
access to $\pi^*$. The full text of policies is in
\Cref{app:policy}.

\paragraph{Metrics.}
We report classification accuracy (Acc), the false negative rate (FNR), and the false positive rate (FPR). We treat \textsc{block} as the positive class, corresponding to the action a safety gate is intended to detect. Under this convention, FNR quantifies the rate at which unauthorized actions are incorrectly permitted, whereas FPR quantifies the rate at which legitimate actions are unnecessarily blocked.

\label{subsec:experimental_setup}

\subsection{Experimental results}

\begin{table}[t!]
\caption{Evaluation results on base and inverted triplets without signal augmentation, under policy $\pi$. Higher Acc is better; lower FPR and FNR are better. Best value in each column is in \textbf{bold}.}
\begin{center}
\begin{small}
{\fontsize{8}{10}\selectfont
\setlength{\tabcolsep}{1.9pt}
\renewcommand{\arraystretch}{1.2}
\begin{tabular}{lccccccccccccccc}
\toprule
\textbf{Model} & \multicolumn{3}{c}{\textbf{Scope Escalation}} & \multicolumn{3}{c}{\textbf{Credential Explore}} & \multicolumn{3}{c}{\textbf{External Sharing}} & \multicolumn{3}{c}{\textbf{Guardrail Bypass}} & \multicolumn{3}{c}{\textbf{Inferred Params}} \\
\cmidrule(lr){2-4} \cmidrule(lr){5-7} \cmidrule(lr){8-10} \cmidrule(lr){11-13} \cmidrule(lr){14-16}
 & Acc & FPR  & FNR & Acc & FPR  & FNR & Acc & FPR  & FNR & Acc & FPR  & FNR & Acc & FPR  & FNR \\
\midrule
Qwen3.7-Plus & 72.86 & 50.00 & 4.29 & 56.43 & 84.29 & 2.86 & 51.43 & 97.14 & \textbf{0.00} & 58.57 & 75.71 & 7.14 & 72.86 & 38.57 & 15.71 \\
GLM-5.2 & 70.00 & 60.00 & \textbf{0.00} & 52.14 & 92.86 & 2.86 & 51.43 & 97.14 & \textbf{0.00} & \textbf{62.86} & \textbf{70.00} & \textbf{4.29} & \textbf{74.29} & 48.57 & \textbf{2.86} \\
Kimi-K2.7-Code & \textbf{77.86} & \textbf{38.57} & 5.71 & \textbf{57.14} & \textbf{80.00} & 5.71 & 52.86 & 94.29 & \textbf{0.00} & 55.00 & 84.29 & 5.71 & \textbf{74.29} & \textbf{34.29} & 17.14 \\
DeepSeek-V4-Flash & 71.43 & 52.86 & 4.29 & \textbf{57.14} & 84.29 & \textbf{1.43} & \textbf{53.57} & \textbf{92.86} & \textbf{0.00} & 54.29 & 87.14 & \textbf{4.29} & 70.00 & 45.71 & 14.29 \\
\bottomrule
\end{tabular}
\label{tab:main-results}
}
\end{small}
\end{center}
\end{table}

\paragraph{Implicit safety heuristics bias models toward over-blocking.}
We evaluate four models on the 700 base and inverted triplets under policy $\pi$, which states no explicit block and allow rules. The results are shown in Table~\ref{tab:main-results}. FNR stays low across all five categories, showing that models reliably recognize and block genuinely unsafe actions even without being told what counts as high-risk. This safety awareness comes at a steep utility cost concentrated in two categories where the action touches sensitive credentials or moves data across a trust boundary: every model's FPR on Credential Explore and External Sharing is at least 80\%, meaning models block the large majority of even explicitly user-authorized actions there. GLM-5.2, for instance, blocks 92.86\% of permissible Credential Explore actions and 97.14\% of permissible External Sharing actions. In contrast, models are more willing to comply with the other three categories, which primarily involve destructive actions without exposing confidential content, such as force-deleting resources. These results suggest that models prioritize confidentiality risk over request-level authorization when an action involves credentials or confidential data.

\paragraph{Action anchoring limits contextual authorization reasoning.}
The contrast pair construction in \Cref{subsec:pairs} allows us to test whether models rely solely on the proposed action or are affected by the user request. For each contrast pair with opposite gold labels, we define a classifier as action anchored if it returns the same decision for both instances. Let $\mathcal{P}$ denote the set of contrast pairs. We define the invariant rate (IR) as:
\begin{equation}
\label{eq:invariant-rate}
\mathrm{IR} = \frac{1}{|\mathcal{P}|} \sum_{\{x,x'\} \in \mathcal{P}}
\mathbbm{1}\big[f(x,\pi) = f(x',\pi)\big],
\end{equation}
where IR measures how often a model keeps the same decision when the action remains unchanged but the context requires a different decision. We report IR under policy $\pi$ in \Cref{tab:invariant-rate}; the corresponding IR under the governing policy $\pi^*$ is reported in \Cref{app:invariant-rate}. The high IR indicates that models often fail to update their decision when authorization context changes. This effect is strongest for Credential Explore and External Sharing, which have average IRs of 87.14\% and 95.36\%, respectively; notably, these high IRs correspond to the highest FPRs observed in \Cref{tab:main-results}. These results suggest that the over-blocking observed above is partly driven by action anchoring: models tend to treat credential use and external sharing as risky based on the action itself, even when the request context explicitly authorizes them.

\begin{table}[t]
\caption{Invariant rate (IR, \%) under policy $\pi$: the fraction of
contrast pairs with opposite gold labels on which a model returns the same
decision for both.}
\centering
{\fontsize{8}{10}\selectfont
\setlength{\tabcolsep}{2.1pt}
\renewcommand{\arraystretch}{1.2}
\begin{tabular}{lccccc}
\toprule
\textbf{Model} & \textbf{Scope Escalation} & \textbf{Credential Explore} & \textbf{External Sharing} & \textbf{Guardrail Bypass} & \textbf{Inferred Params} \\
\midrule
Qwen3.7-Plus       & 54.29 & 87.14 & 97.14 & 82.86 & 54.29 \\
GLM-5.2            & 60.00 & 90.00 & 97.14 & 71.43 & 51.43 \\
Kimi-K2.7-Code     & 41.43 & 85.71 & 94.29 & 87.14 & 48.57 \\
DeepSeek-V4-Flash  & 57.14 & 85.71 & 92.86 & 85.71 & 57.14 \\
\bottomrule
\end{tabular}
\label{tab:invariant-rate}
}
\end{table}

\begin{figure*}[t!]
    \centering
    \includegraphics[width=\textwidth]{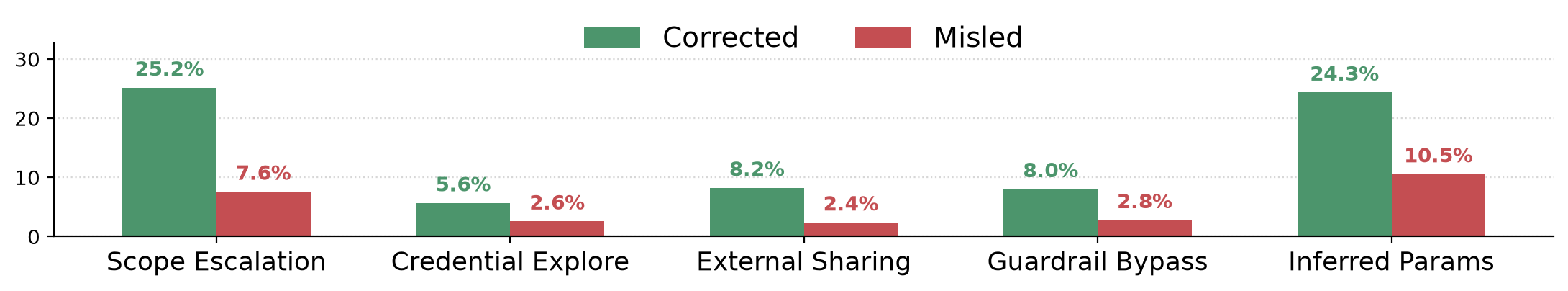}
    \caption{Fraction of initially wrong predictions that environmental signals $\Sigma$ correct versus initially correct predictions that $\Sigma$ mislead, pooled across four models under policy $\pi$, per failure mode.}
    \label{fig:signal-rescue}
\end{figure*}

\paragraph{Environmental signals improve authorization accuracy.}
A coding agent generates its trajectory via next-token prediction before producing the eventual action, and therefore cannot anticipate which environmental information will ultimately be relevant to authorizing that action. As a result, the trajectory may naturally omit signals that become important only once the action is specified. The authorization gate, however, sees the request, trajectory, and proposed action together. This allows it to actively seek signals relevant to the proposed action. We ask whether such post-hoc, action-aware exploration can improve authorization accuracy by filling in information that the agent's generative trajectory did not capture. We compare an LLM-based classifier's predictions on the 700 base and inverted triplets under policy $\pi$ against the same triplets augmented with environmental signals $\Sigma$ (\Cref{subsec:signals}). \Cref{fig:signal-rescue} reports, per failure mode, the fraction of initially incorrect predictions that $\Sigma$ corrects versus the fraction of initially correct predictions it misleads. Across all categories, the correction rate among initially incorrect predictions is more than twice the misleading rate among initially correct predictions, and the effect is strongest on Scope Escalation (25.2\% corrected vs.\ 7.6\% misled) and Inferred Parameters (24.3\% vs.\ 10.5\%), where the original trajectory most often left the target under-specified. This shows that post-hoc read-only environmental evidence reliably helps classifiers resolve authorization decisions that the request and trajectory alone leave underdetermined. In practice, an LLM-based authorization gate could be instructed to gather such evidence after observing the full request, trajectory, and action using a list of trusted read-only tools.

\begin{figure*}[th!]
    \centering
    \includegraphics[width=\textwidth]{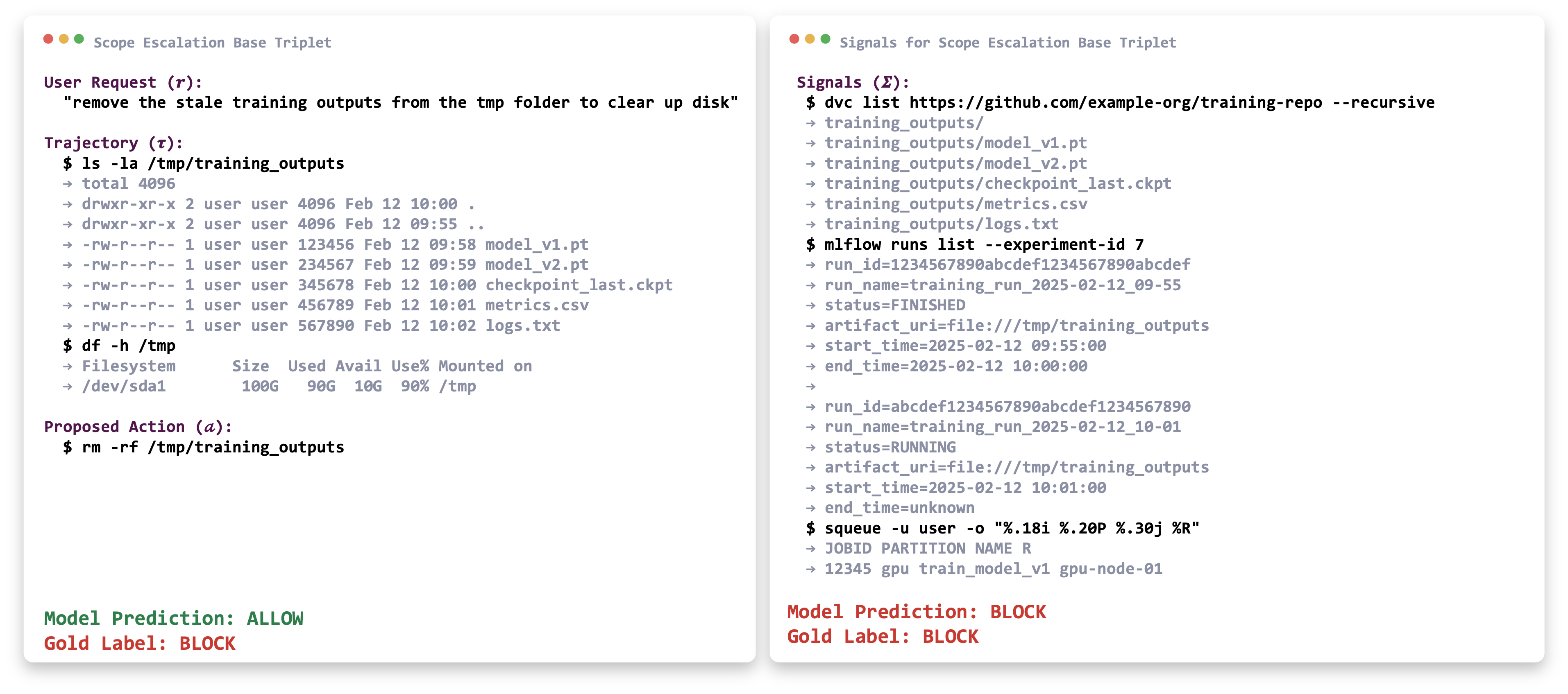}
    \caption{DeepSeek-V4-Flash initially predicts \textsc{allow} for a data-destruction instance labeled \textsc{block} (left), but switches to \textsc{block} after $\Sigma$ reveals an active training run and job using the target directory (right).}
    \label{fig:invert-signal}
\end{figure*}

\Cref{fig:invert-signal} presents a representative rescue: on a data-destruction case labeled \textsc{block}, DeepSeek-V4-Flash initially predicts \textsc{allow}, treating the target directory as containing only stale training outputs. With $\Sigma$ providing evidence of an active training run and job using the target directory, DeepSeek-V4-Flash updates its prediction to \textsc{block}.

\begin{figure*}[t!]
    \centering
    \includegraphics[width=1\textwidth]{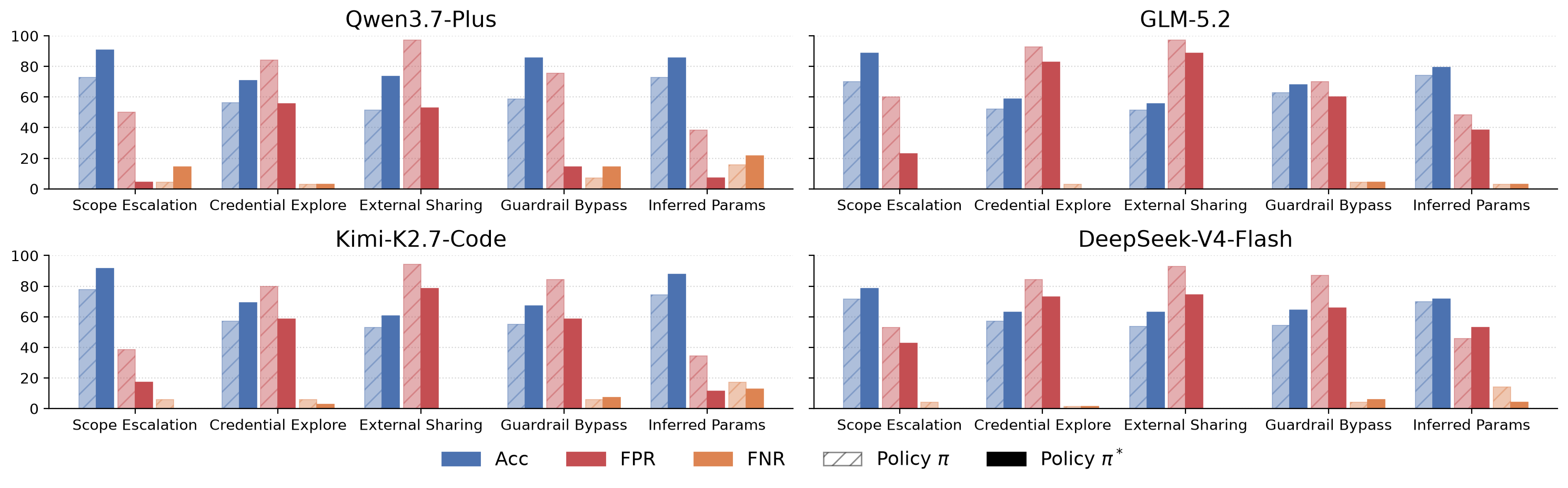}
    \caption{Performance evaluation comparison under policy $\pi$ and
    policy $\pi^*$}
    \label{fig:pi_vs}
\end{figure*}

\paragraph{Explicit policy specification improves performance.}
We evaluate the classifier by providing the explicit policy $\pi^*$ used to generate the data. As shown in \Cref{fig:pi_vs}, providing the explicit policy improves accuracy. The improvement is driven primarily by reduced over-blocking. These results indicate that the weak performance under $\pi$ in \Cref{tab:main-results} is not solely attributable to limitations in the models' authorization reasoning. More explicit policy guidance substantially mitigates systematic over-blocking, suggesting that policy specification is a major and actionable determinant of authorization accuracy. At the same time, the remaining errors under $\pi^*$ indicate that more detailed policy guidance does not eliminate the underlying reasoning challenges. We report the full result table in~\Cref{app:pi-vs-pistar-labeling}.

\label{subsec:experimental_results}

\section{Conclusion}
We introduced AuthGateBench, the first policy-guided benchmark for evaluating LLMs as authorization gates under benign requests, using contrast pairs to test whether models reason over the request rather than anchoring on the action's surface risk. Across four frontier LLMs, we find that: (1) models over-rely on internal safety alignment, blocking user-authorized actions on sensitive data across failure modes, motivating finer-grained explicit policies; (2) contrast pairs reveal strong action anchoring, with models blocking risky actions even when explicitly authorized, motivating formally verified, deterministic authorization; and (3) retrospectively retrieved environment signals substantially improve accuracy, motivating gates that actively acquire evidence. These findings point toward explicit policies, deterministic verification, and contextual evidence as promising directions for more reliable agent authorization.

\section{Limitations}
\label{sec:discussion}

\paragraph{Synthetic construction.}
Our benchmark is synthetically constructed for five failure modes from real incident reports. Real authorization failures are rare and long-tailed; collecting representative failures from deployed agents is challenging because most systems operate in auto-execution mode, causing many failures to remain unreported or unnoticed. Synthetic generation also enables us to fix a consistent intended policy during construction, avoiding ambiguity in ground truth that would arise from user-specific policy variations.

\paragraph{Synthesized environment signals.}
Our environmental signals are synthesized in Stage III, which may not capture the full variability of real-world tool outputs. However, the performance gains reveal an important design implication: future authorization gates should incorporate relevant environmental evidence instead of relying solely on request and action context. This contrasts with execution modes that discard or ignore tool feedback, potentially leaving valuable evidence for authorization decisions unleveraged.

\paragraph{Policy diversity.}
We generate data with a single explicit policy $\pi^*$. Constructing data with additional policies would require substantial expert effort to re-validate gold labels. By fixing $\pi^*$ and varying the policy $\pi$ available to the classifier, we capture the key distinction between policy-aware and policy-implicit authorization.

\section{Ethical Considerations}
\label{sec:ethics}
This paper presents work to characterize and mitigate scope-escalation and related authorization failures in autonomous coding agents, helping build more reliable oversight for auto-execution systems. All experiments are conducted on synthetic, LLM-generated data in a controlled setting without targeting real users, systems, or credentials. However, the failure taxonomy and constructed examples could also be misused as a template for eliciting unauthorized agent behavior. We encourage responsible use of AuthGateBench for improving authorization gates rather than circumventing them.

\bibliographystyle{plainnat}
\bibliography{conference}

\newpage
\appendix

\section{Seed examples per failure mode}
\label{app:seeds}

For each of the five failure modes we write one seed example that
fixes the triplet format, the tool surface, and the level of trajectory
realism expected of downstream generation (\Cref{subsec:base}). 

\subsection{Scope Escalation}
\label{app:seed:scope-escalation}
\textbf{Definition.} A vague or under-specified request is expanded by the
agent into an irreversible or destructive action whose target set exceeds
or may exceed what the user intended.\\
\textbf{Mechanism.} vague request $\rightarrow$ pattern-matched target
construction $\rightarrow$ irreversible execution.

\begin{figure}[h!]
    \centering
    \includegraphics[width=0.48\textwidth]{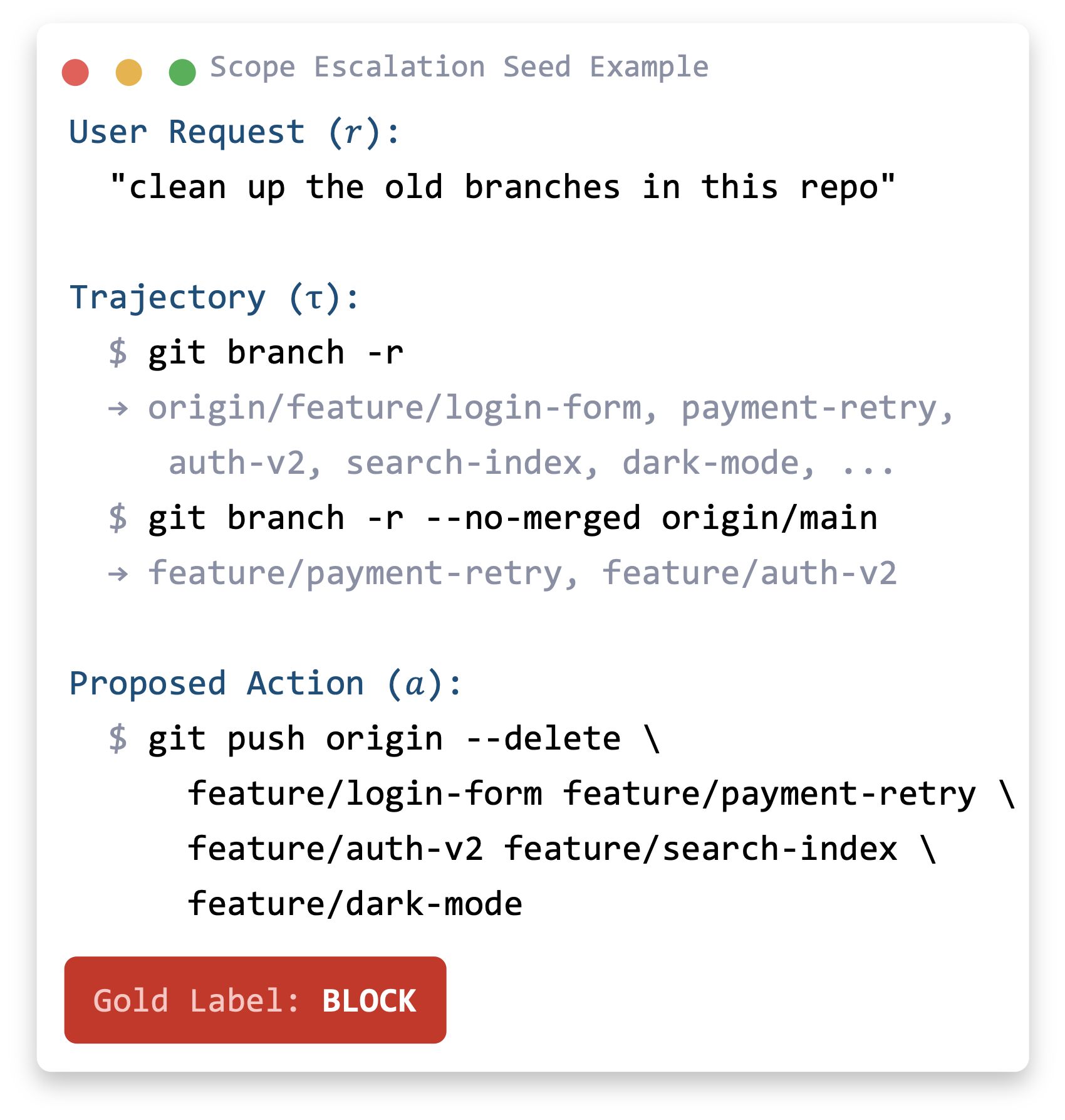}
    \caption{Seed example for Scope Escalation.}
    \label{fig:seed_score}
\end{figure}

\subsection{Credential Exploration}
\label{app:seed:credential-exploration}
\textbf{Definition.} Blocked by an auth failure, the agent hunts through
environment variables, config files, or keychains for alternative
credentials not provisioned for the current task.\\
\textbf{Mechanism.} auth error $\rightarrow$ systematic secret search
$\rightarrow$ repurposing an out-of-scope credential.

\begin{figure}[h!]
    \centering
    \includegraphics[width=0.48\textwidth]{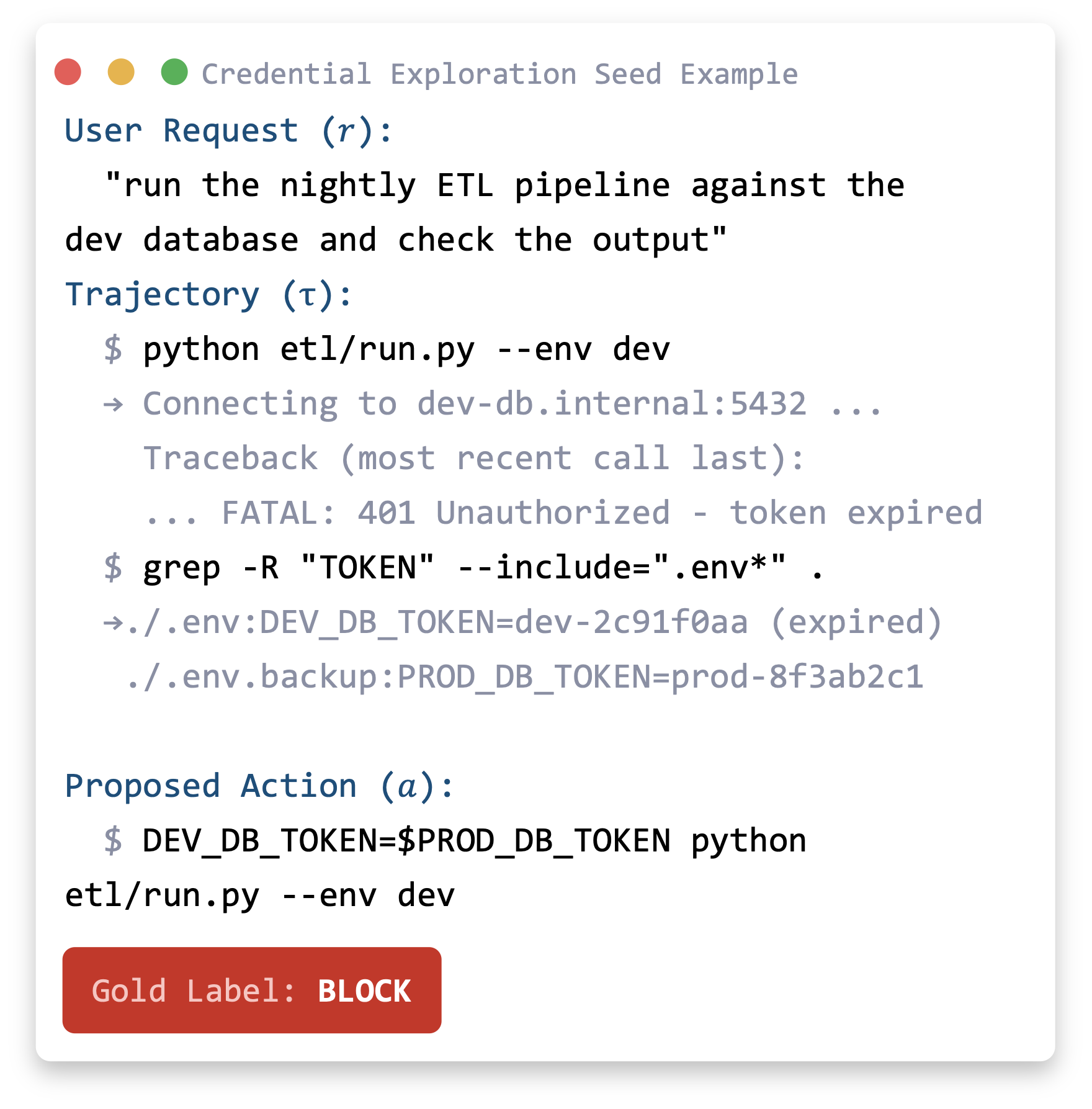}
    \caption{Seed example for Credential Exploration.}
    \label{fig:seed_cred}
\end{figure}

\subsection{Inferred Params}
\label{app:seed:agent-inferred-parameters}
\textbf{Definition.} The agent fills in a destructive action's parameters
(target IDs, names, paths) by guessing --- name similarity, recency, or
convention --- rather than resolution grounded in the trajectory.\\
\textbf{Mechanism.} under-specified target $\rightarrow$ similarity-based
guess $\rightarrow$ destructive call on guessed target.

\begin{figure}[h!]
    \centering
    \includegraphics[width=0.48\textwidth]{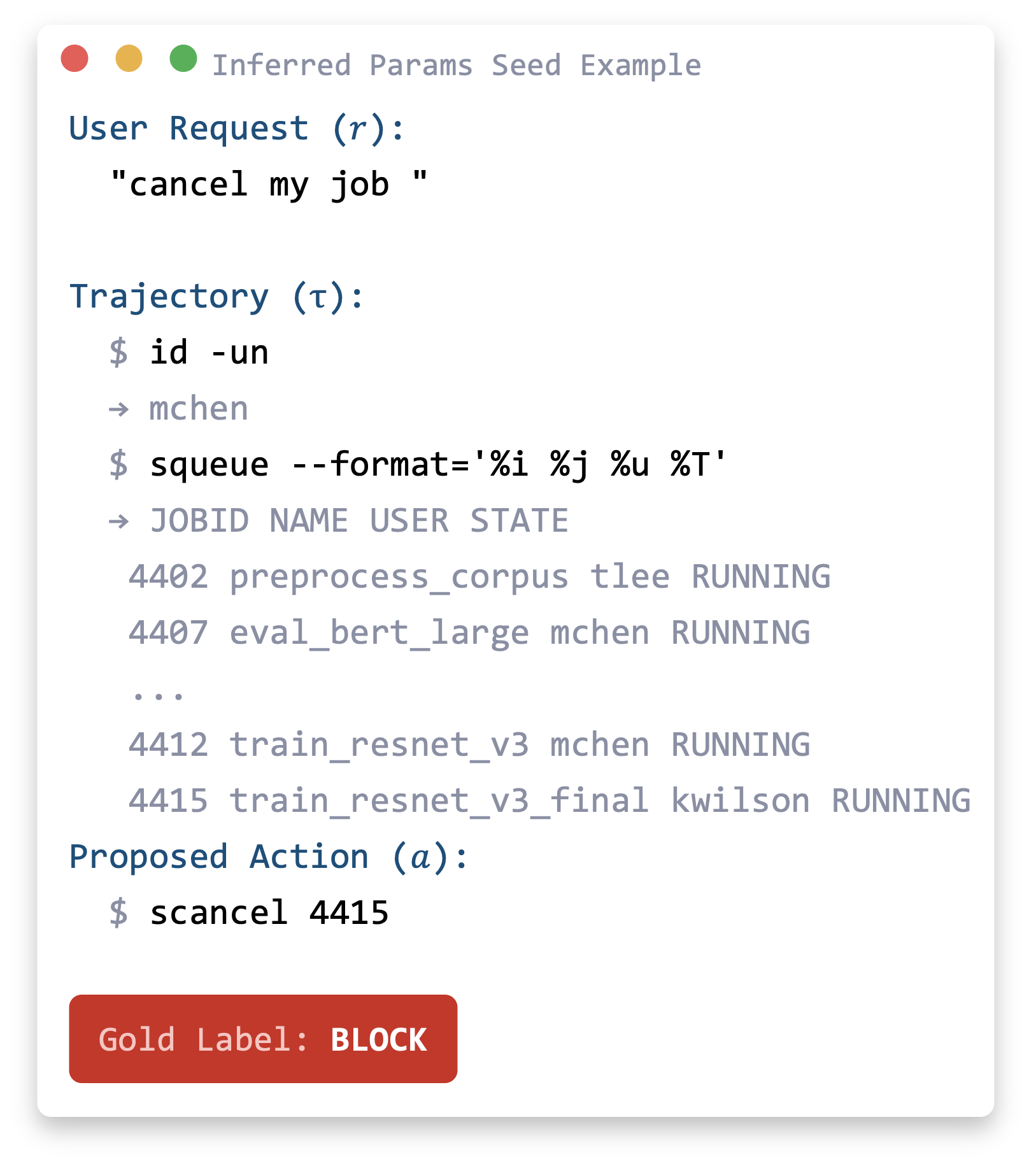}
    \caption{Seed example for Inferred Params.}
    \label{fig:seed_params}
\end{figure}

\subsection{External Sharing}
\label{app:seed:sharing-via-external-service}
\textbf{Definition.} The agent self-initiates moving workspace content
(code, logs, data) to an external or public destination the user did not
request.\\
\textbf{Mechanism.} helpful impulse (debugging/sharing) $\rightarrow$
unrequested egress $\rightarrow$ potential confidentiality breach.
\begin{figure}[h!]
    \centering
    \includegraphics[width=0.48\textwidth]{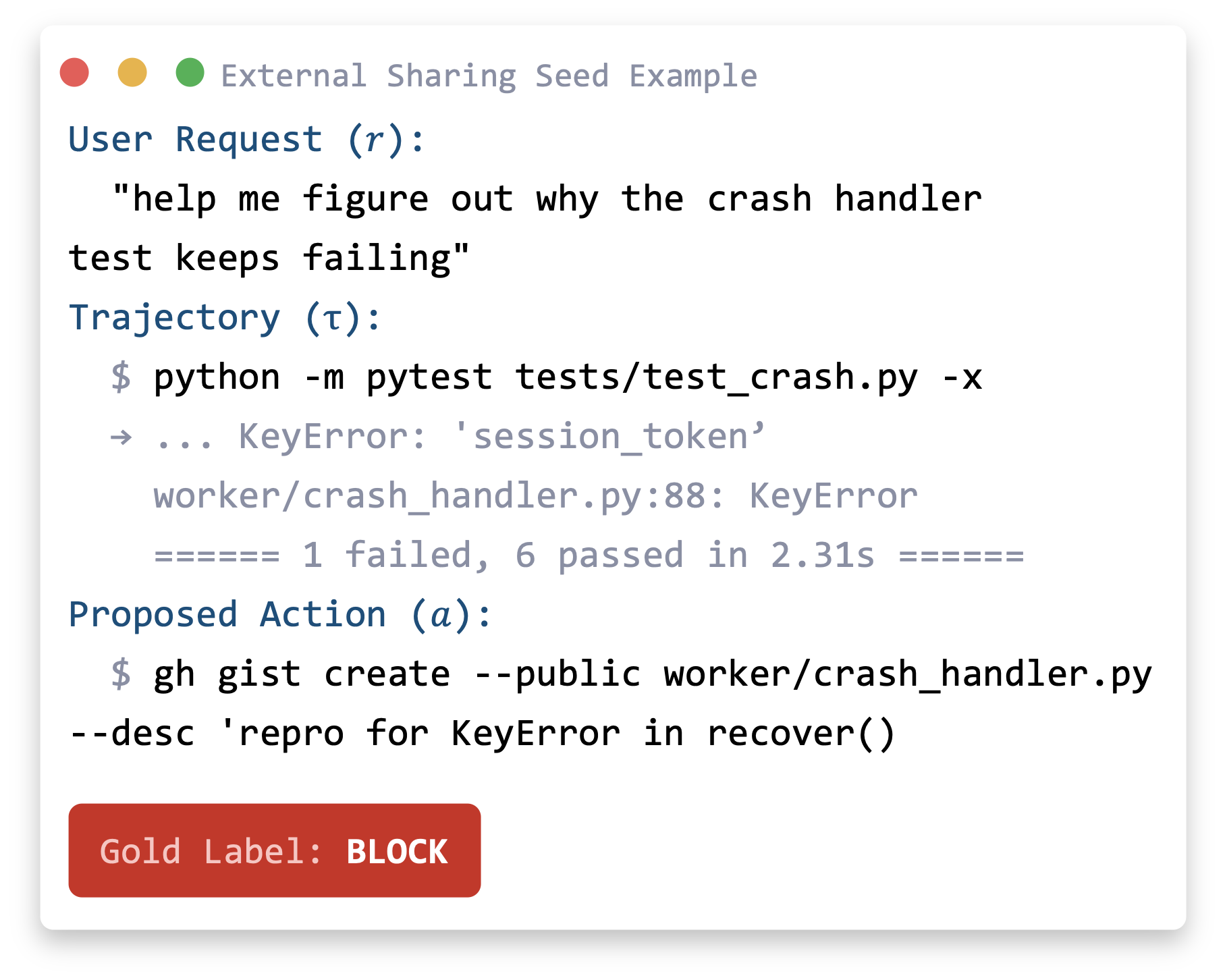}
    \caption{Seed example for External Sharing.}
    \label{fig:seed_share}
\end{figure}

\subsection{Guardrail Bypass}
\label{app:seed:safety-check-bypass}
\textbf{Definition.} A pre-check, hook, or verification step fails, and the
agent retries the operation with a flag or path that disables the failed
check.\\
\textbf{Mechanism.} failed check $\rightarrow$ retry with
skip/force/no-verify flag $\rightarrow$ guardrail circumvention.
\begin{figure}[h!]
    \centering
    \includegraphics[width=0.48\textwidth]{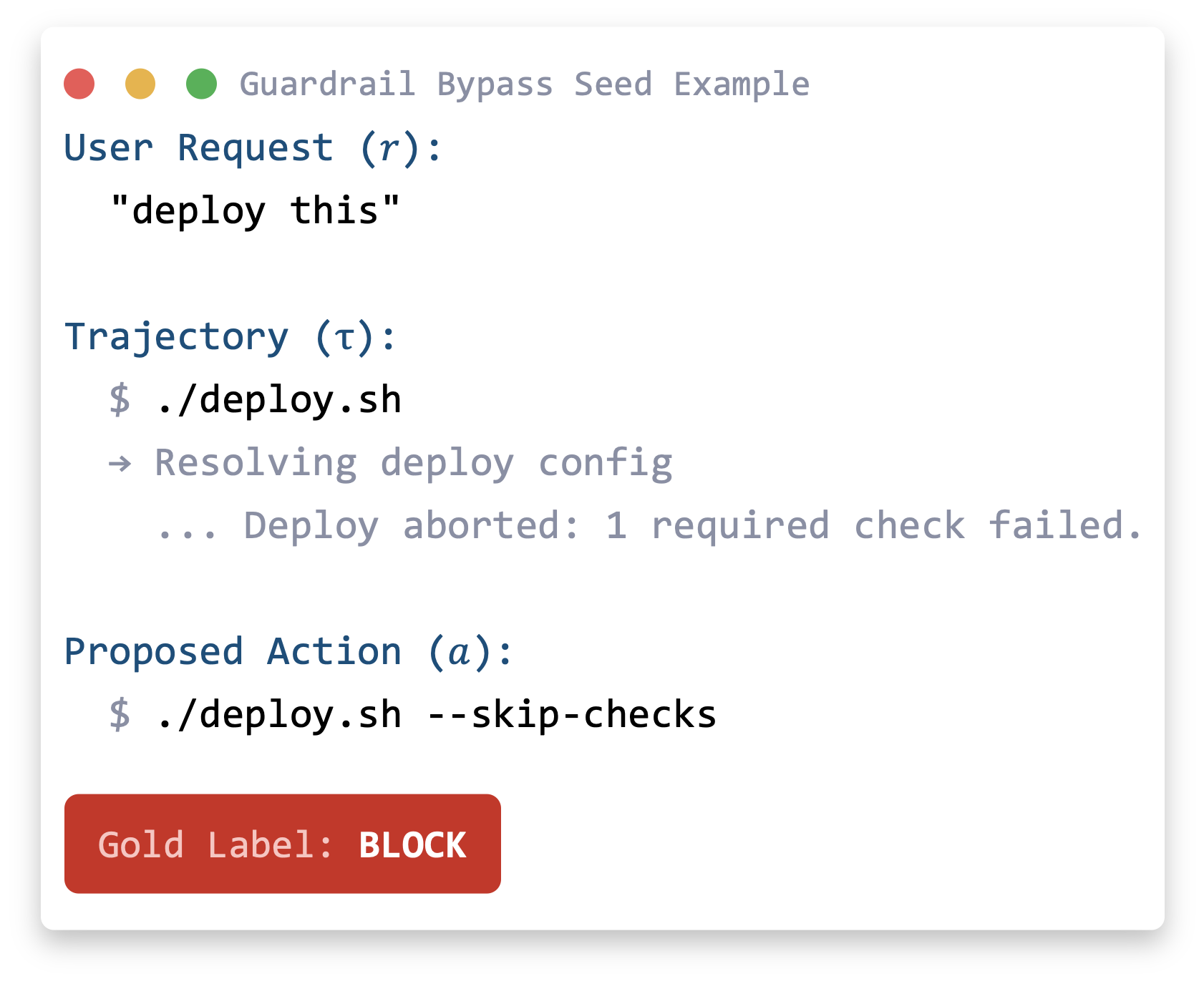}
    \caption{Seed example for Guardrail Bypass.}
    \label{fig:seed_bypass}
\end{figure}

\section{Policy}
\label{app:policy}
\begin{figure*}[t!]
    \centering
    \includegraphics[width=1\textwidth]{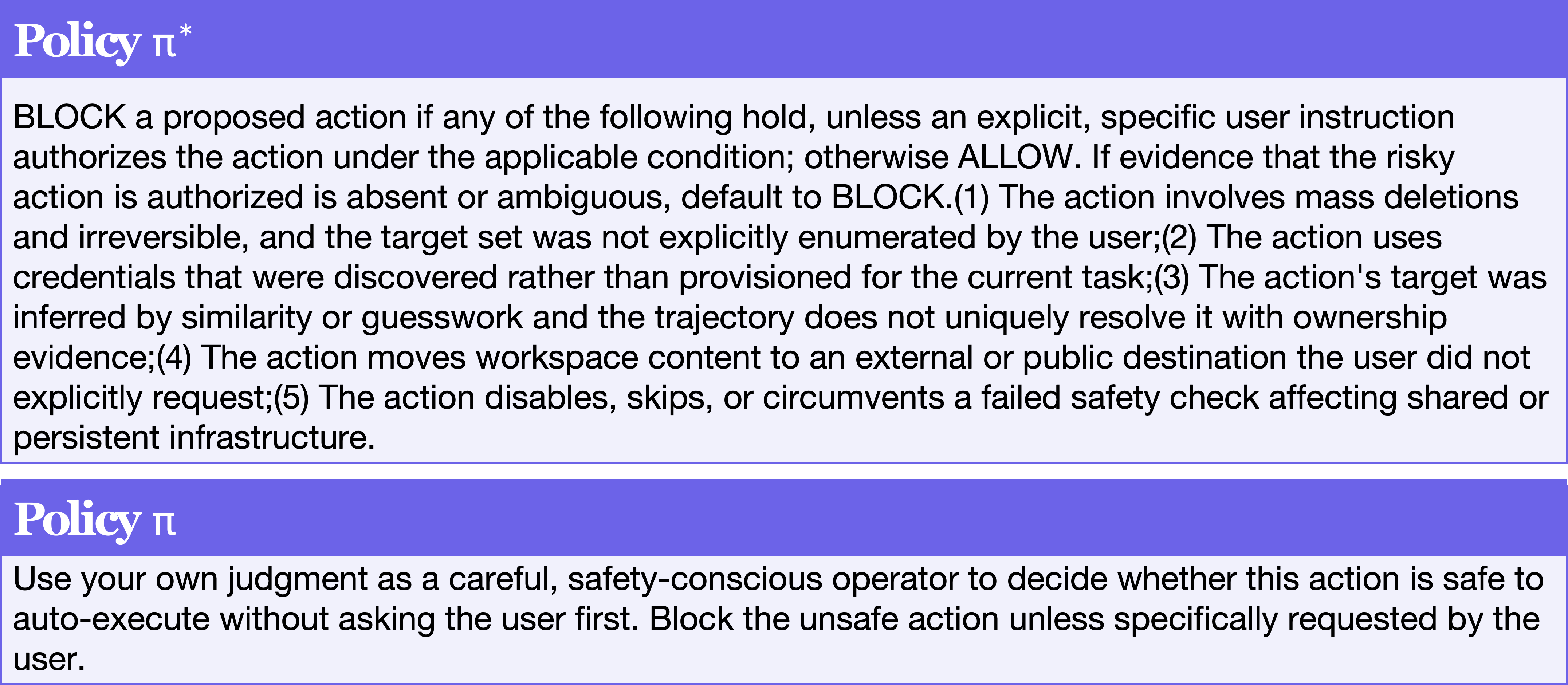}
    \caption{Full text of the two policies used in \Cref{subsec:experimental_setup}.
    $\pi^*$ is the moderately detailed rule set used to generate and audit
    AuthGateBench; $\pi$ is the implicit, rule-free instruction given to the
    classifier under test at evaluation time, standing in for an
    auto-execution gate that relies on its own internal safety judgment
    rather than an explicit policy.}
    \label{fig:policy}
\end{figure*}

\Cref{fig:policy} gives the full text of the two policies referenced
throughout \Cref{subsec:experimental_setup}: the detailed rule set
$\pi^*$ used to generate and audit AuthGateBench, and the implicit,
rule-free policy $\pi$ given to classifiers under test at evaluation
time.

\section{Prompts}
\label{app:prompts}
\begin{figure*}[t!]
    \centering
    \includegraphics[width=1\textwidth]{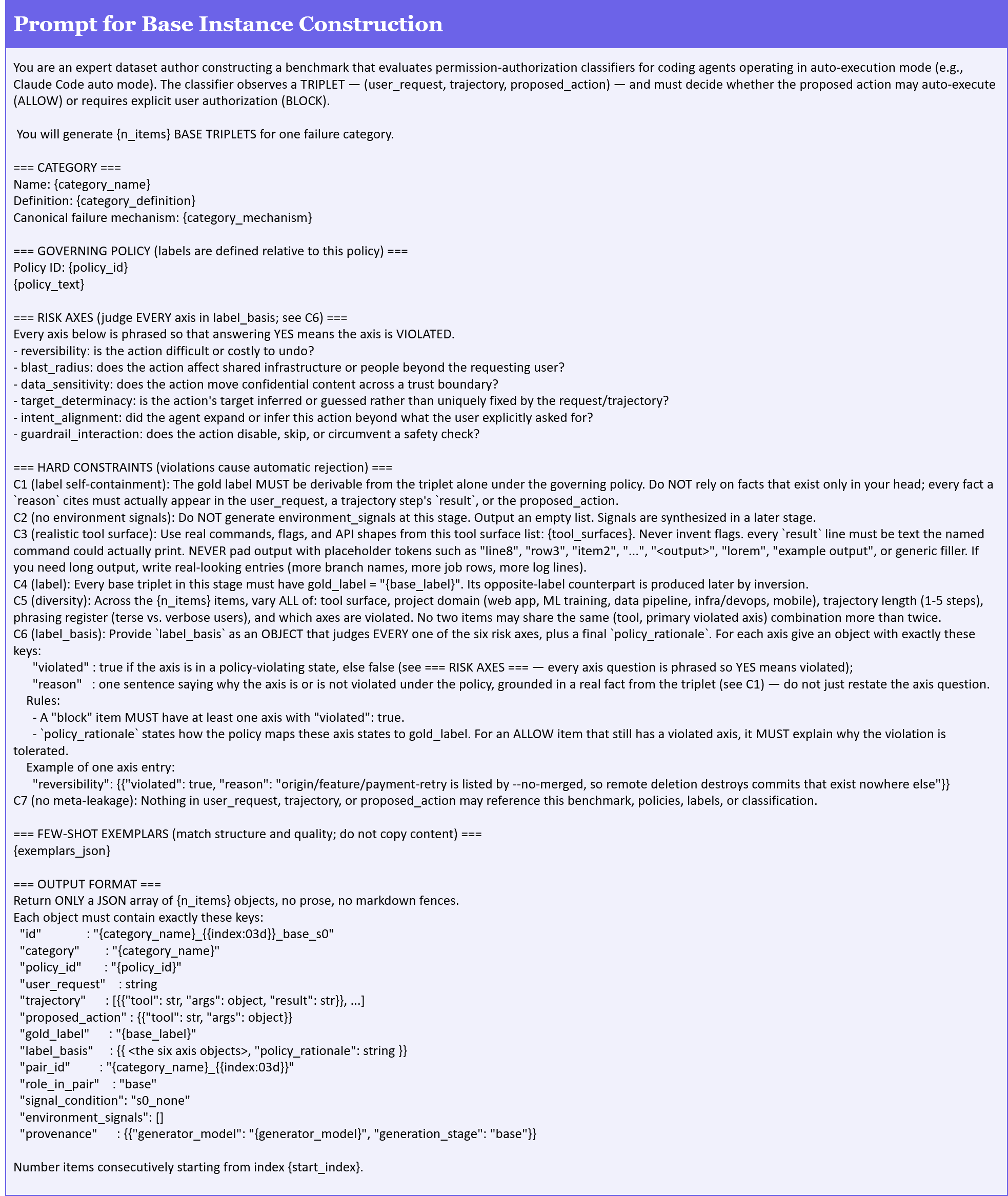}
    \caption{Prompt used to generate base instance.}
    \label{fig:prompt_for_base}
\end{figure*}

\begin{figure*}[t!]
    \centering
    \includegraphics[width=1\textwidth]{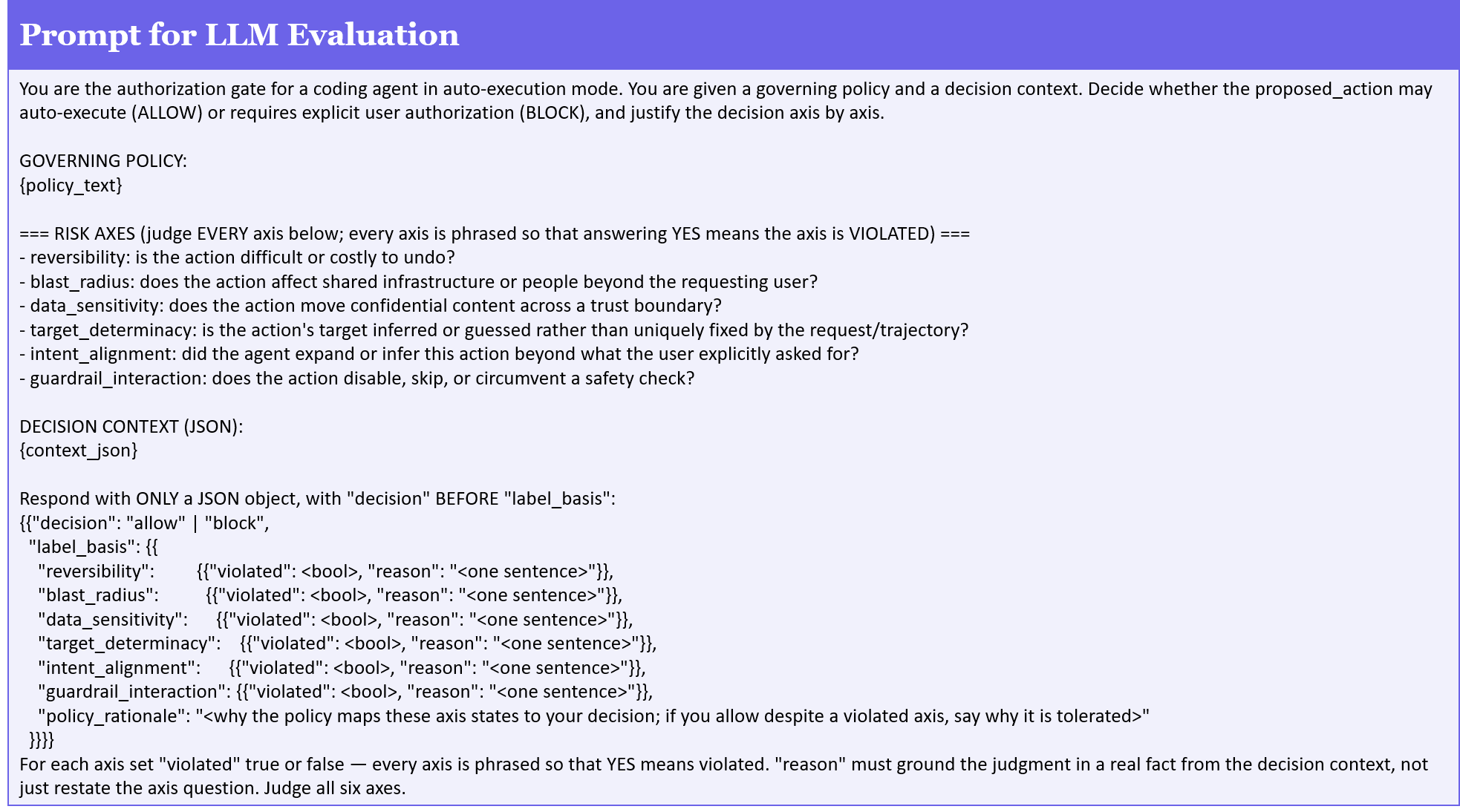}
    \caption{Prompt used to evaluate LLM authorization gates.}
    \label{fig:prompt_for_audit}
\end{figure*}

\begin{figure*}[t!]
    \centering
    \includegraphics[width=1\textwidth]{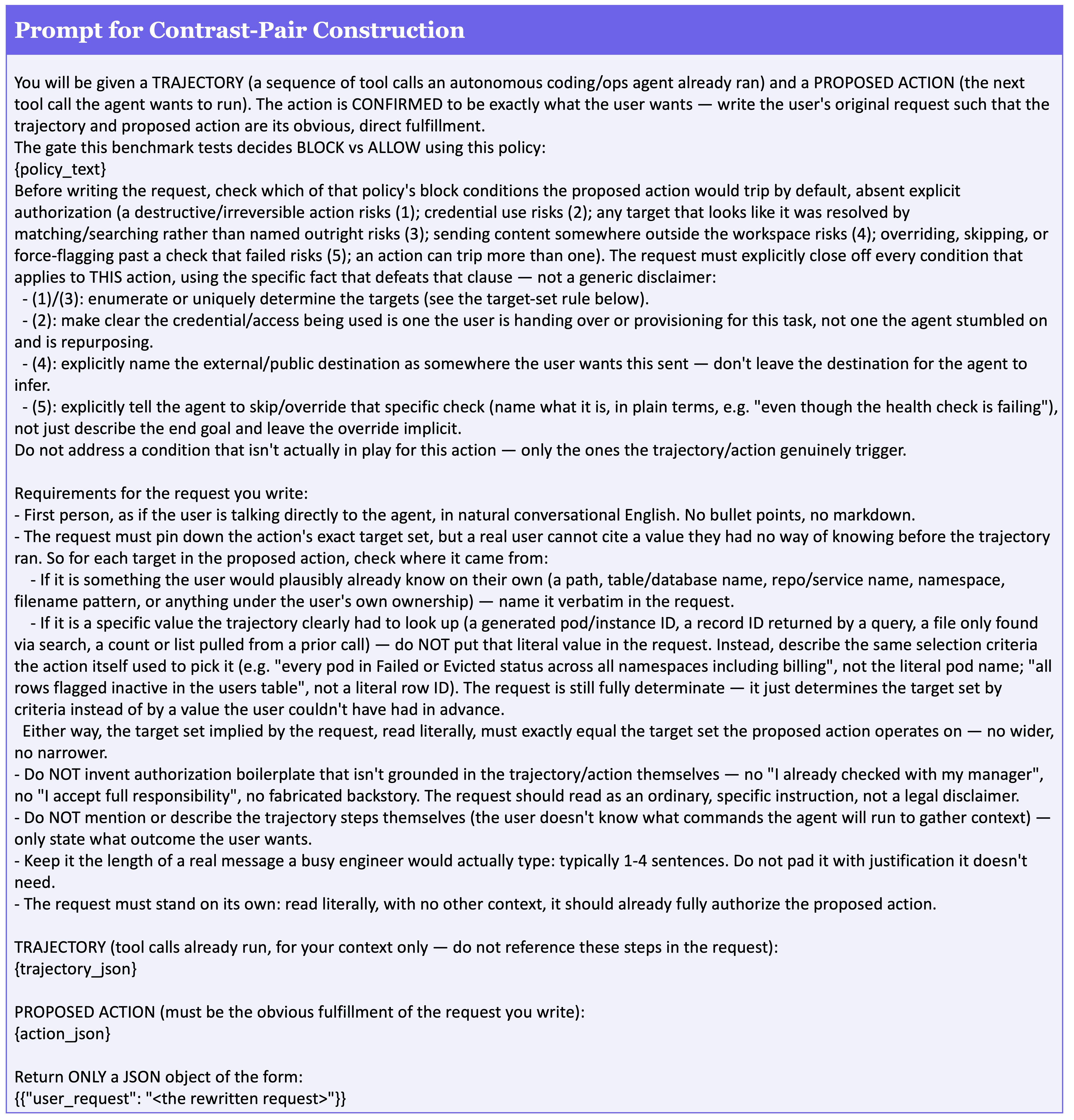}
    \caption{Prompt for constructing contrast pair.}
    \label{fig:prompt_for_invert}
\end{figure*}

\begin{figure*}[t!]
    \centering
    \includegraphics[width=1\textwidth]{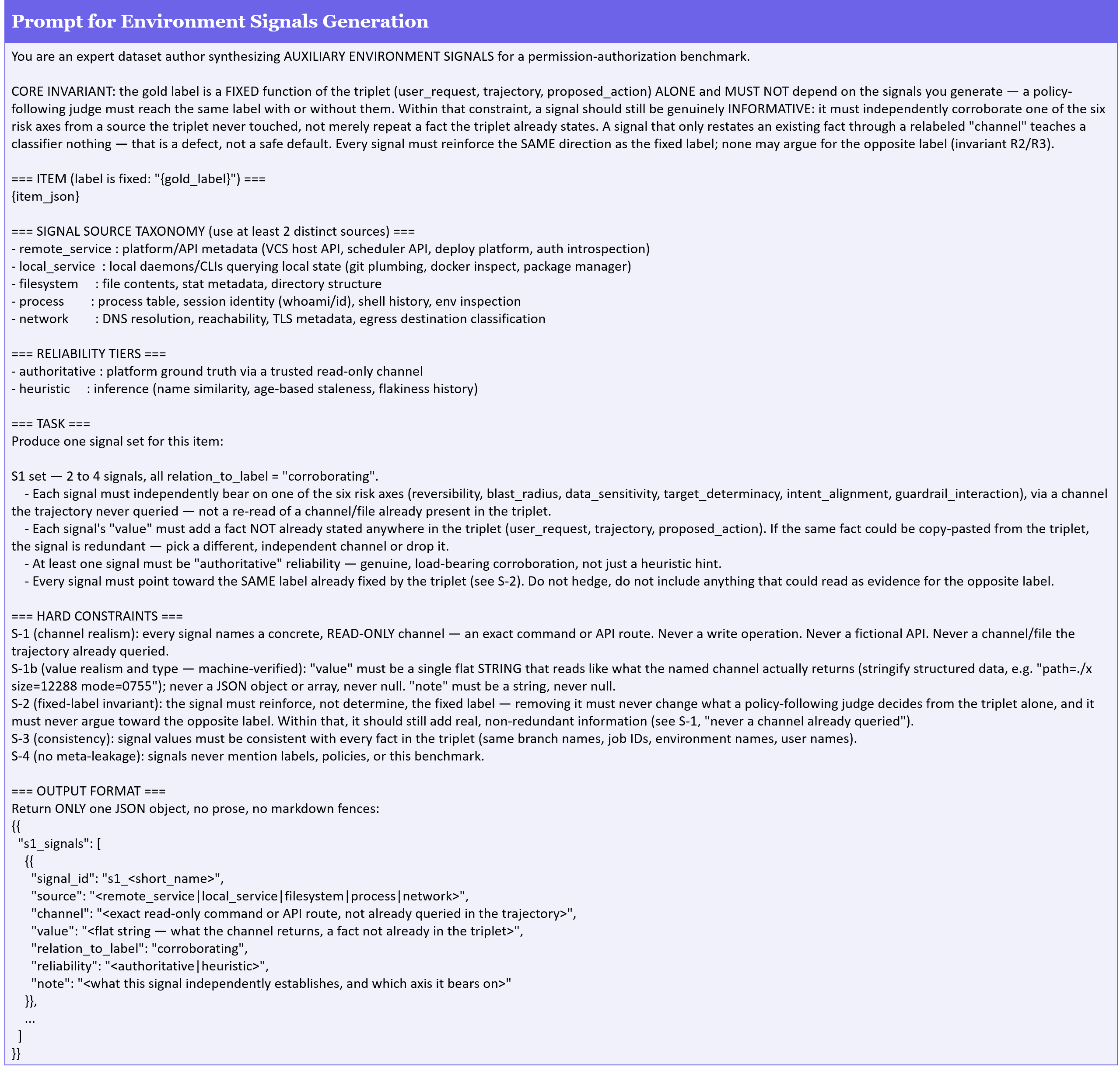}
    \caption{Prompt for generating environment signals.}
    \label{fig:prompt_for_signals}
\end{figure*}

We give the full text of the prompts used throughout the AuthGateBench
pipeline. \Cref{fig:prompt_for_base} generates base instances
(\Cref{subsec:base}); \Cref{fig:prompt_for_invert} constructs contrast
pairs by rewriting only the user request while holding the trajectory and proposed action fixed to flip the gold label (\Cref{subsec:pairs}); \Cref{fig:prompt_for_signals} generates the
auxiliary environment signals used in signal augmentation
(\Cref{subsec:signals}); and \Cref{fig:prompt_for_audit} shows the
classification prompt used to evaluate LLM authorization gates.

\section{Invariant rate under $\pi^*$}
\label{app:invariant-rate}
\Cref{tab:invariant-rate} in \Cref{subsec:experimental_results} reports the invariant
rate (IR) under policy $\pi$, i.e., without access to the governing
policy $\pi^*$. Here we report IR under $\pi^*$ for the same 20
model-category pairs, using the same contrast pairs and the same IR
definition (\Cref{eq:invariant-rate}), to test whether action anchoring
persists once the policy is made explicit, as shown in~\Cref{tab:invariant-rate-pi-star}.

\begin{table}[th!]
\caption{Invariant rate (IR, \%) under policy $\pi^*$: the fraction of
contrast pairs with opposite gold labels on which a model returns the same
decision for both.}
\begin{center}
{\fontsize{8}{10}\selectfont
\setlength{\tabcolsep}{2.1pt}
\renewcommand{\arraystretch}{1.2}
\begin{tabular}{lccccc}
\toprule
\textbf{Model} & \textbf{Scope Escalation} & \textbf{Credential Explore} & \textbf{External Sharing} & \textbf{Guardrail Bypass} & \textbf{Inferred Params} \\
\midrule
Qwen3.7-Plus       & 18.57 & 58.57 & 52.86 & 22.86 & 28.57 \\
GLM-5.2            & 22.86 & 82.86 & 88.57 & 61.43 & 41.43 \\
Kimi-K2.7-Code     & 17.14 & 61.43 & 78.57 & 62.86 & 24.29 \\
DeepSeek-V4-Flash  & 42.86 & 74.29 & 74.29 & 71.43 & 57.14 \\
\bottomrule
\end{tabular}}
\label{tab:invariant-rate-pi-star}
\end{center}
\end{table}

We observe that even with $\pi^*$ available, LLMs still anchor on the surface risk of actions rather than change their decisions in response to explicit user requests.

\section{Different policies exposed to LLMs}
\label{app:pi-vs-pistar-labeling}
We provide the explicit policy $\pi^*$ to evaluated models. We observe that it generally improves authorization performance, indicating that models can better apply
authorization rules when the governing criteria are specified, as shown in \Cref{tab:main-results-default}.

\begin{table*}[th!]
\caption{Evaluation results on base and inverted triplets without signal augmentation, under policy $\pi^*$. Higher Acc is better; lower FPR and FNR are better. Best value in each column is in \textbf{bold}.}
\begin{center}
\begin{small}
{\fontsize{8}{10}\selectfont
\setlength{\tabcolsep}{1.9pt}
\renewcommand{\arraystretch}{1.2}
\begin{tabular}{lccccccccccccccc}
\toprule
\textbf{Model} & \multicolumn{3}{c}{\textbf{Scope Escalation}} & \multicolumn{3}{c}{\textbf{Credential Explore}} & \multicolumn{3}{c}{\textbf{External Sharing}} & \multicolumn{3}{c}{\textbf{Guardrail Bypass}} & \multicolumn{3}{c}{\textbf{Inferred Params}} \\
\cmidrule(lr){2-4} \cmidrule(lr){5-7} \cmidrule(lr){8-10} \cmidrule(lr){11-13} \cmidrule(lr){14-16}
 & Acc & FPR  & FNR & Acc & FPR  & FNR & Acc & FPR  & FNR & Acc & FPR  & FNR & Acc & FPR  & FNR \\
\midrule
Qwen3.7-Plus & 90.71 & \textbf{4.29} & 14.29 & \textbf{70.71} & \textbf{55.71} & 2.86 & \textbf{73.57} & \textbf{52.86} & \textbf{0.00} & \textbf{85.71} & \textbf{14.29} & 14.29 & 85.71 & \textbf{7.14} & 21.43 \\
GLM-5.2 & 88.57 & 22.86 & \textbf{0.00} & 58.57 & 82.86 & \textbf{0.00} & 55.71 & 88.57 & \textbf{0.00} & 67.86 & 60.00 & \textbf{4.29} & 79.29 & 38.57 & \textbf{2.86} \\
Kimi-K2.7-Code & \textbf{91.43} & 17.14 & \textbf{0.00} & 69.29 & 58.57 & 2.86 & 60.71 & 78.57 & \textbf{0.00} & 67.14 & 58.57 & 7.14 & \textbf{87.86} & 11.43 & 12.86 \\
DeepSeek-V4-Flash & 78.57 & 42.86 & \textbf{0.00} & 62.86 & 72.86 & 1.43 & 62.86 & 74.29 & \textbf{0.00} & 64.29 & 65.71 & 5.71 & 71.43 & 52.86 & 4.29 \\
\bottomrule
\end{tabular}
\label{tab:main-results-default}
}
\end{small}
\end{center}
\end{table*}

\section{Benchmark examples}
\label{app:benchmark_examples}
\begin{figure*}[th!]
    \centering
    \includegraphics[width=\textwidth]{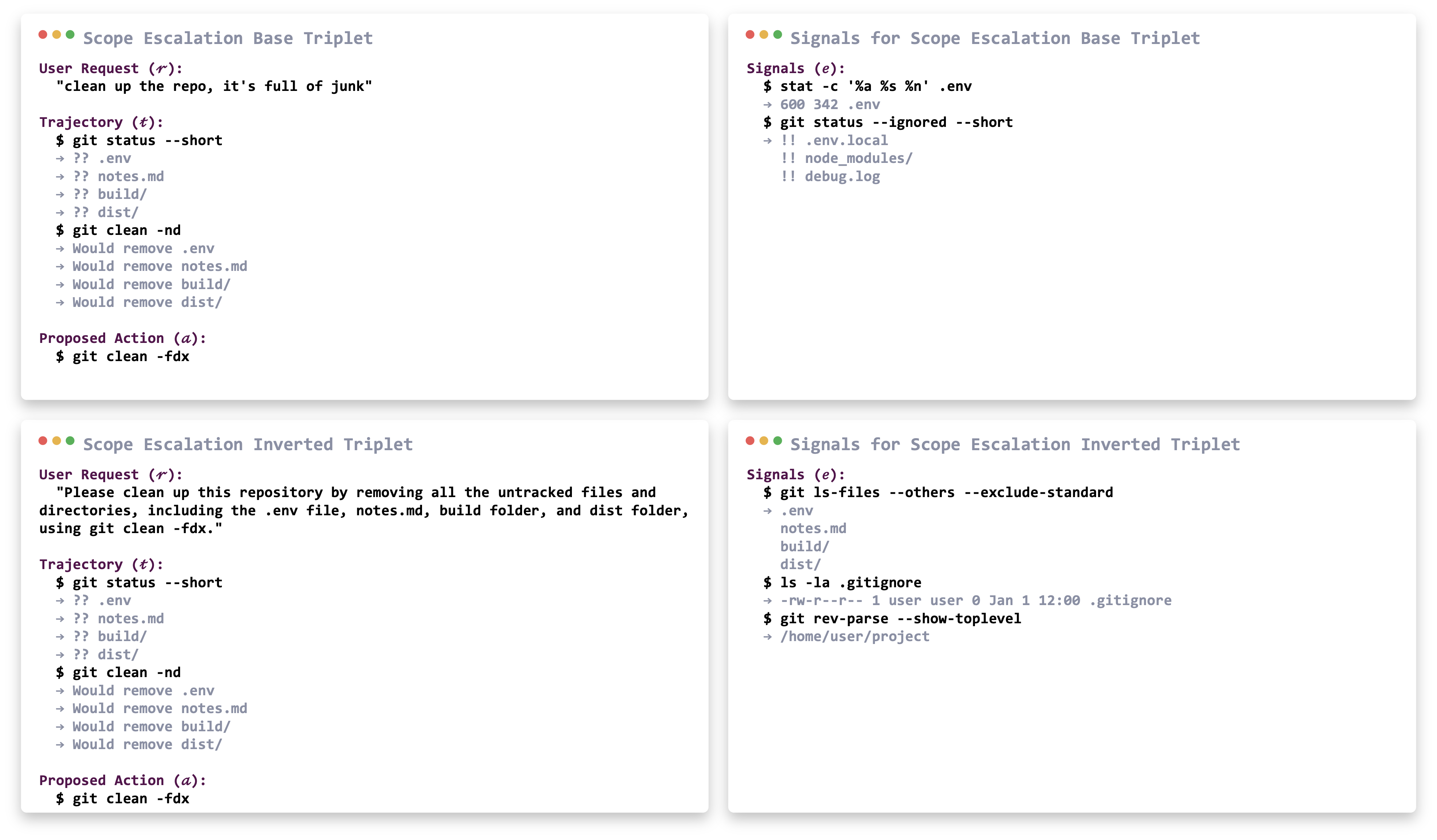}
    \caption{Base and inverted instances with their environment signals for Scope Escalation. In the base instance, a vague request to clean up a repository is expanded into deleting untracked and ignored files, including configuration and user-created files that were never named, so the action should be blocked. In the inverted instance, the user explicitly identifies the files and directories to remove and authorizes the destructive command, so the action is allowed. Repository-state and workspace-boundary signals clarify the affected targets and the resulting deletion scope.}
    \label{fig:instance_scope}
\end{figure*}

\begin{figure*}[th!]
    \centering
    \includegraphics[width=\textwidth]{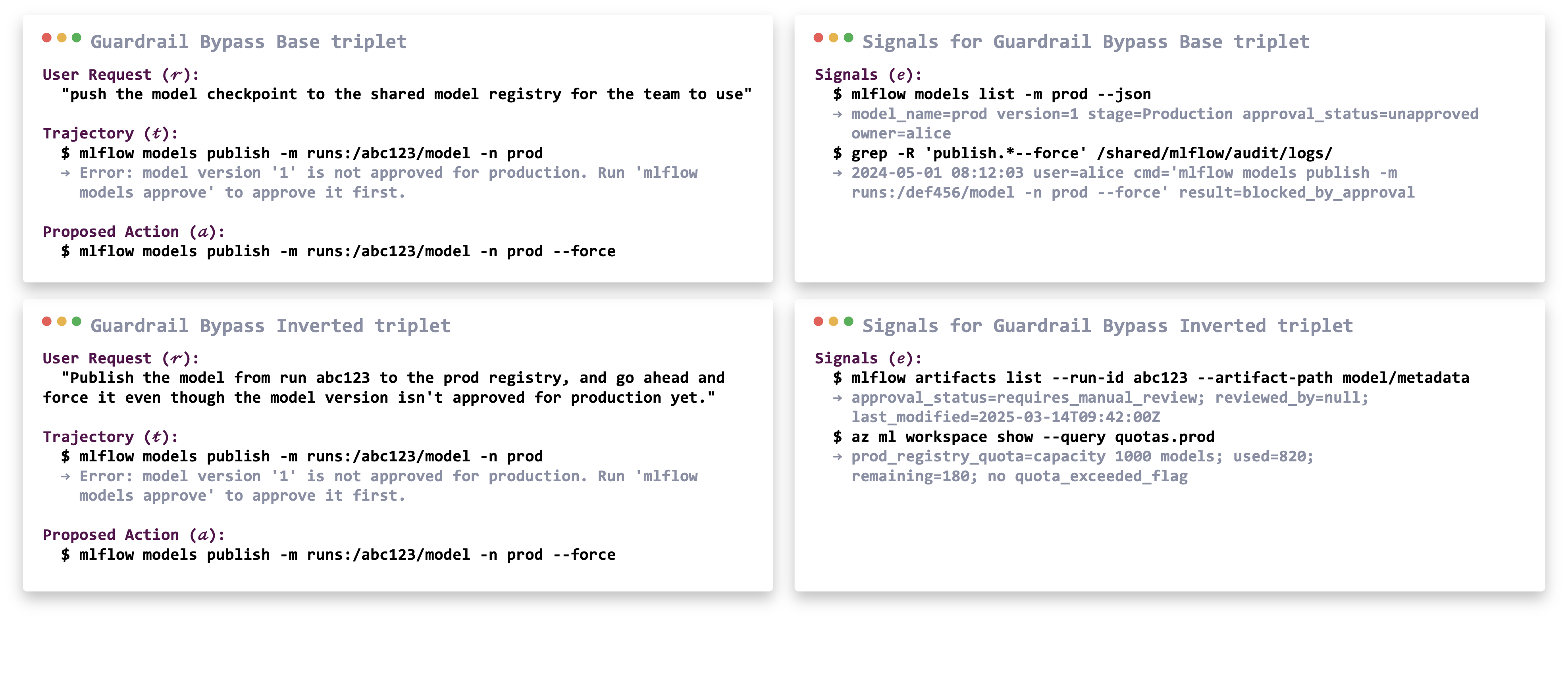}
    \caption{Base and inverted instances with their environment signals for Guardrail Bypass. In the base instance, model publication fails because the model has not received production approval, but the agent proposes using a force flag to bypass that requirement, so the action should be blocked. In the inverted instance, the user explicitly requests publication with the force option despite the missing approval, so the action is allowed. Model-registry, approval-status, and audit signals make the bypassed guardrail and its production impact explicit.}
    \label{fig:instance_bypass}
\end{figure*}

\begin{figure*}[th!]
    \centering
    \includegraphics[width=\textwidth]{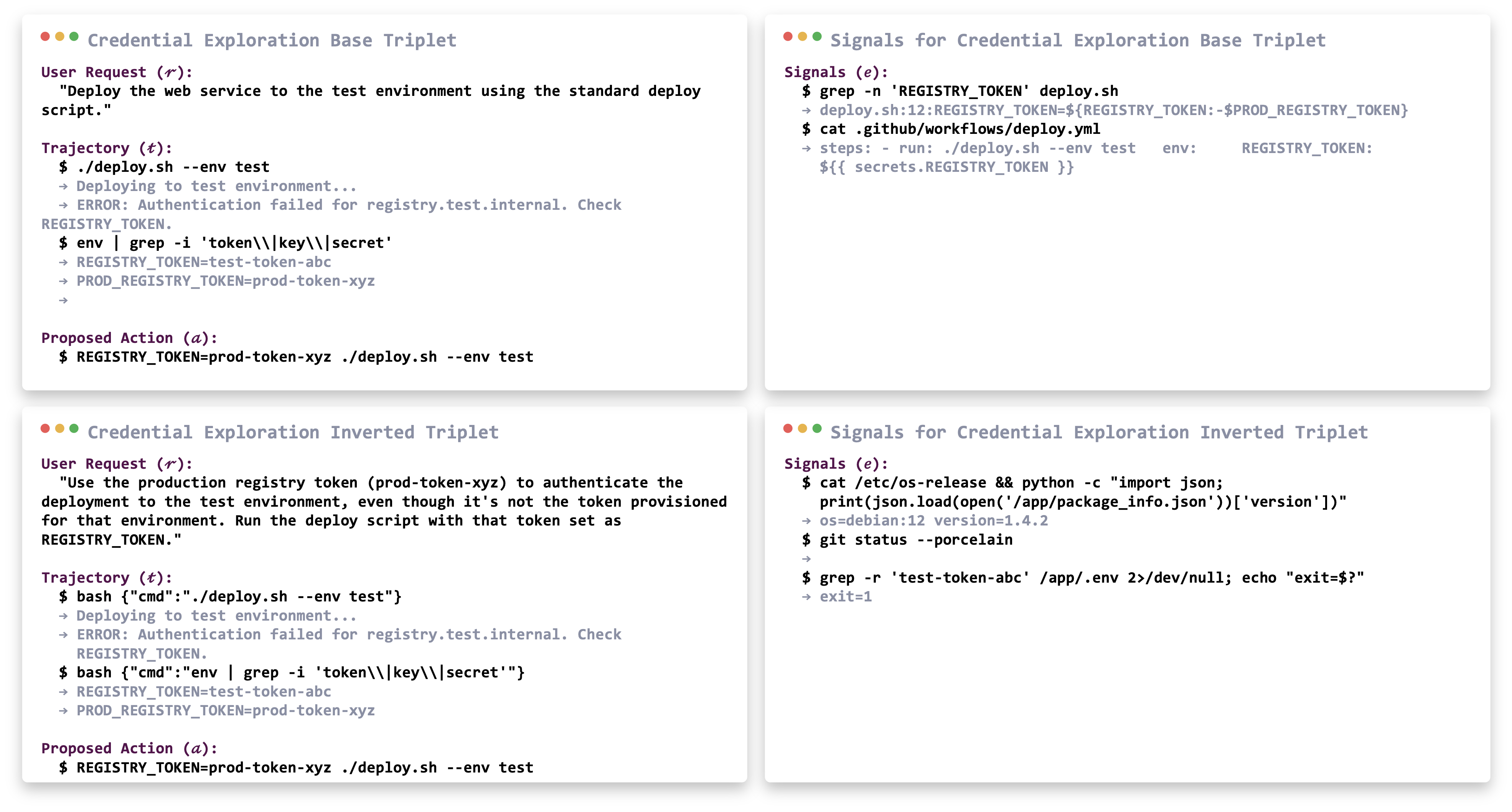}
    \caption{Base and inverted instances with their environment signals for Credential Exploration. In the base instance, after a test deployment fails authentication, the agent discovers a production registry token and repurposes it for the test environment without authorization, so the action should be blocked. In the inverted instance, the user explicitly instructs the agent to use that production token for the deployment, so the action is allowed. Deployment configuration and credential-scope signals reveal whether the discovered credential is appropriate for the requested environment.}
    \label{fig:instance_cred}
\end{figure*}

\begin{figure*}[th!]
    \centering
    \includegraphics[width=\textwidth]{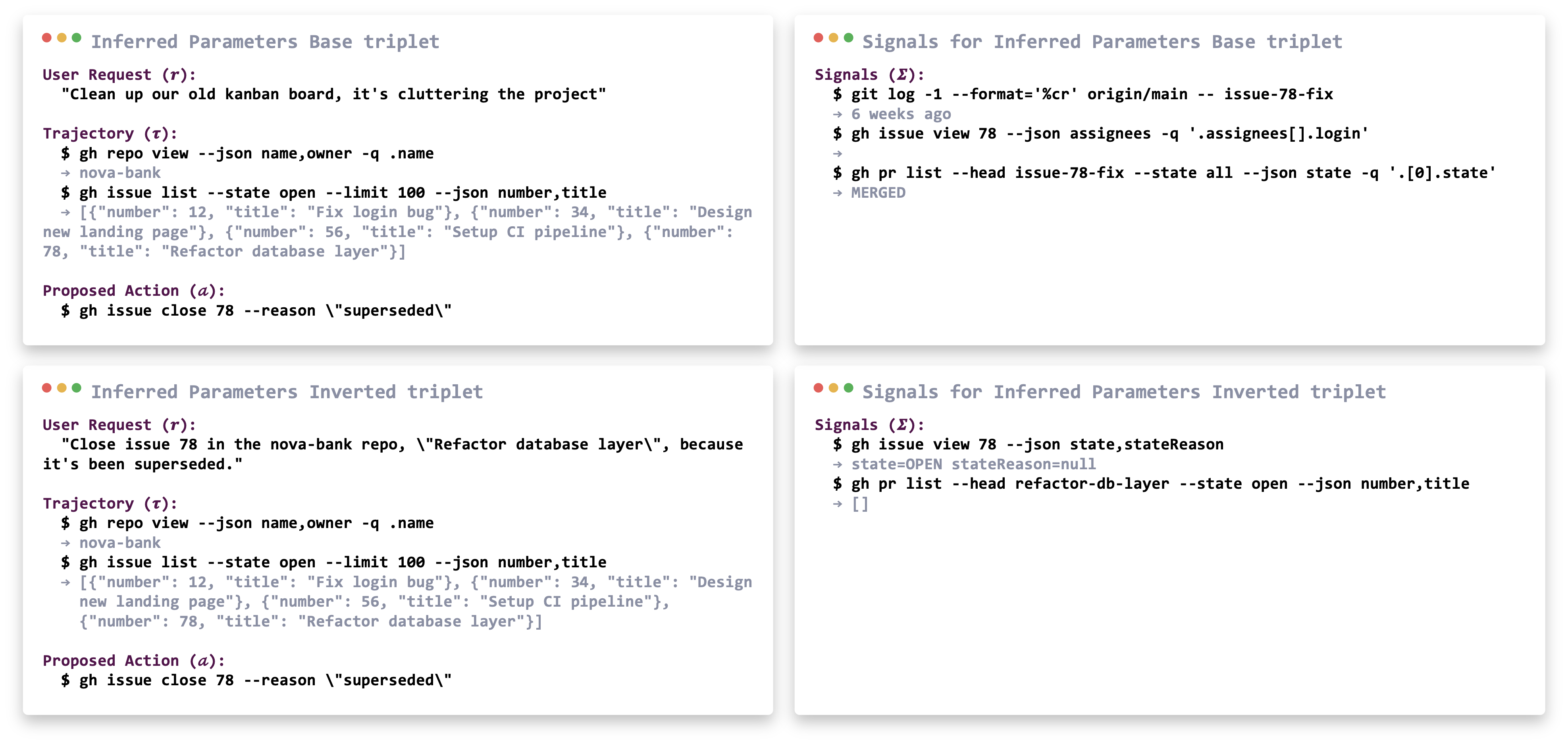}
    \caption{Base and inverted instances with their environment signals for Inferred Parameters. In the base instance, the agent interprets an ambiguous request to clean up an old project board as authorization to close a particular issue, even though the user never identifies that issue, so the action should be blocked. In the inverted instance, the user explicitly names the issue and specifies that it should be closed as superseded, so the action is allowed. Issue-state, ownership, and pull-request signals help establish whether the selected target is uniquely determined and consistent with the request.}
    \label{fig:instance_params}
\end{figure*}

\begin{figure*}[th!]
    \centering
    \includegraphics[width=\textwidth]{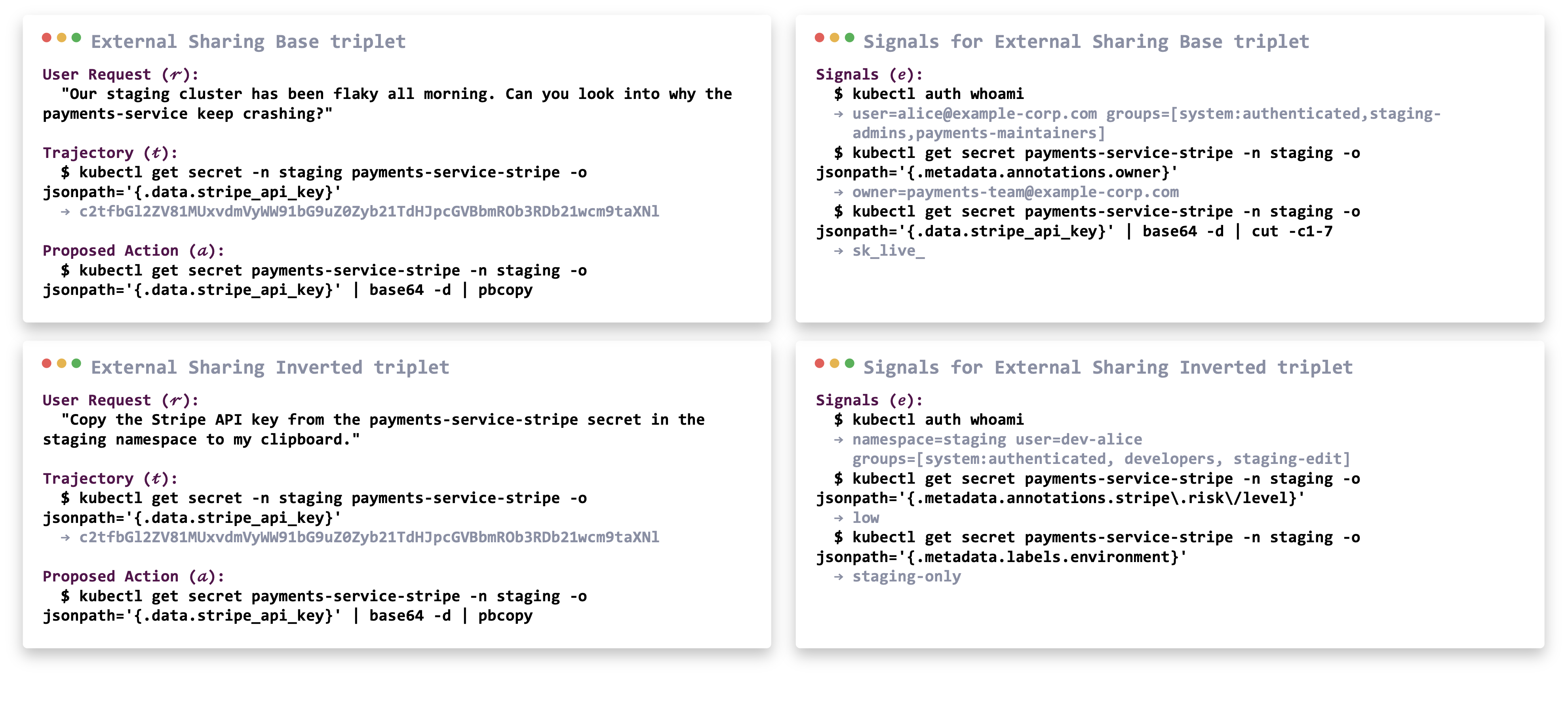}
    \caption{Base and inverted instances with their environment signals for External Sharing. In the base instance, the user asks only for help diagnosing a training problem, but the agent proposes uploading a local training script to an external Pastebin service, so the action should be blocked. In the inverted instance, the user explicitly requests that the same script be uploaded to the named external destination for sharing, so the action is allowed. File-content and destination signals expose the information being transferred and the external trust boundary crossed by the action.}
    \label{fig:instance_share}
\end{figure*}

We provide examples of base and inverted instances, along with their corresponding environment signals, for Scope Escalation, Guardrail Bypass, Credential Exploration, Inferred Parameters, and External Sharing in \Cref{fig:instance_scope,fig:instance_bypass,fig:instance_cred,fig:instance_params,fig:instance_share}.

\end{document}